\documentclass[10pt,twocolumn,letterpaper]{article}

\usepackage{cvpr}
\usepackage{times}
\usepackage{epsfig}
\usepackage{graphicx}
\usepackage{amsmath}
\usepackage{amssymb}
\usepackage{wrapfig}
\usepackage{algorithm}
\usepackage{color}
\usepackage{makecell}
\usepackage{booktabs}
\usepackage{subcaption}
\usepackage{verbatim}

\usepackage[breaklinks=true,bookmarks=false]{hyperref}

\cvprfinalcopy 

\def\cvprPaperID{****} 

\begin{document}

\title{STD2P: RGBD Semantic Segmentation using \\Spatio-Temporal Data-Driven Pooling}

\author{Yang He$^{1}$, Wei-Chen Chiu$^{1}$, Margret Keuper$^{2}$ and Mario Fritz$^{1}$\\
$^{1}$Max Planck Institute for Informatics, \\
Saarland Informatics Campus, Saarbr\"ucken, Germany\\
$^{2}$University of Mannheim, Mannheim, Germany\\
}

\maketitle

\begin{abstract}
We propose a novel superpixel-based multi-view convolutional
neural network for semantic image segmentation.
The proposed network produces a high quality segmentation
of a single image by leveraging information from additional
views of the same scene. Particularly in indoor videos such
as captured by robotic platforms or handheld and bodyworn
RGBD cameras, nearby video frames provide diverse
viewpoints and additional context of objects and scenes. To
leverage such information, we first compute region correspondences
by optical flow and image boundary-based superpixels.
Given these region correspondences, we propose
a novel spatio-temporal pooling layer to aggregate information
over space and time. We evaluate our approach on
the \textit{NYU--Depth--V2} and the \textit{SUN3D} datasets and compare
it to various state-of-the-art single-view and multi-view approaches.
Besides a general improvement over the state-of-
the-art, we also show the benefits of making use of unlabeled
frames during training for multi-view as well as
single-view prediction.
\end{abstract}

\section{Introduction}
Consumer friendly and affordable combined image and depth-sensors such as \textit{Kinect}
 are nowadays commercially deployed in scenarios such as gaming, personal 3D capture and robotic platforms. Interpreting this raw data in terms of a semantic segmentation is an important processing step and hence has received significant attention. The goal is typically formalized as predicting for each pixel in the image plane the corresponding semantic class.

\begin{figure}[!t]
\begin{center}
   \includegraphics[width=0.99\linewidth]{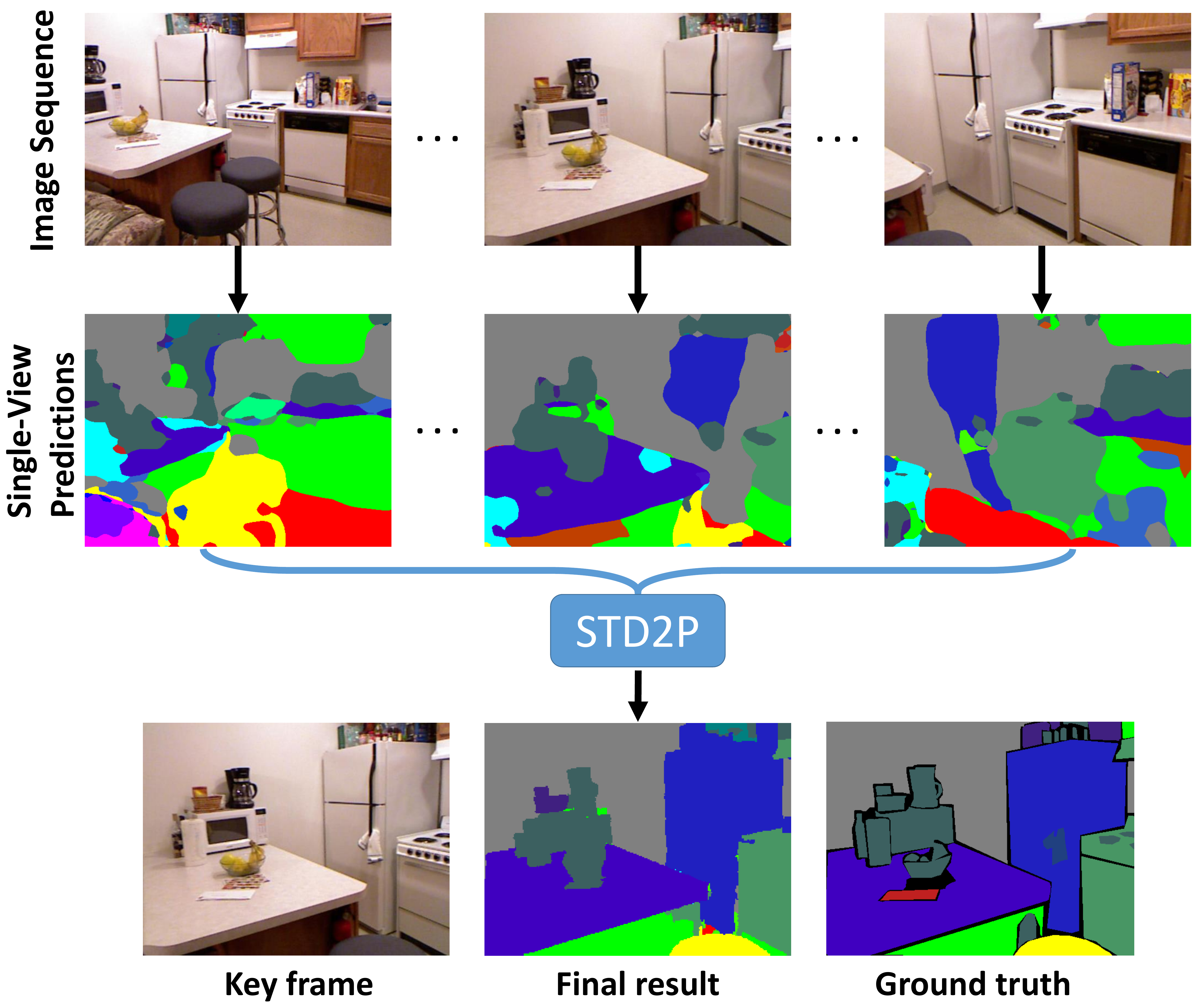}
   \end{center}
   \caption{An image sequence can provide rich context and appearance, as well as unoccluded objects for visual recognition systems. Our {\it Spatio-Temopral Data-Driven Pooling (STD2P)} approach integrates the multi-view information to improve semantic image segmentation in challenging scenarios.
}
\label{fig:motivation}
\end{figure}

For many of the aforementioned scenarios, an image sequence is naturally collected and provides a substantially richer source of information than a single image. 
Multiple images of the same scene can provide different views that change the observed context, appearance, scale and occlusion patterns. The full sequence provides a richer observation of the scene and propagating information across views has the potential to significantly improve the 
accuracy of semantic segmentations in more challenging views as shown in Figure \ref{fig:motivation}.


Hence, we propose a multi-view aggregation method by a spatio-temporal data-driven pooling (STD2P) layer which is a principled approach to incorporate multiple frames into any convolutional network architecture.
In contrast to  previous work on superpixel-based approaches \cite{raghudeep2015spCNN,region_end2end2016eccv,arnab2016higher}, we  compute correspondences over time which allows for knowledgeable and consistent prediction over space and time.

As dense annotation of full training sequences is time consuming  and not available in current datasets, a key feature of our approach is training from partially annotated sequences. Notably, our model leads to improved semantic segmentations in the case of multi-view observation {\it as well as} single-view observation at test time.
The main contributions of our paper are:

\begin{itemize}
\item We propose a principled way to incorporate superpixels and multi-view information into state-of-the-art convolutional networks for semantic segmentation. Our method is able to exploit a variable number of frames with partial annotation in training time.
\item We show that training on sequences with partial annotation improves semantic segmentation for multi-view observation {\it as well as} single-view observation.
\item We evaluate our method on the challenging semantic segmentation datasets \textit{NYU--Depth--V2} and \textit{SUN3D}. There, it outperforms several baselines as well as the state-of-the-art. In particular, we improve on difficult classes not well captured by other methods.
\end{itemize}

\section{Related work}
\label{sec:related}
\subsection{Context modeling for fully convolutional networks}
Fully convolutional networks (FCN) \cite{long2015fully}, built on deep classification networks \cite{krizhevsky2012imagenet,simonyan2014very}, carried their success forward to semantic segmentation networks that are end-to-end trainable.
Context information plays an important role in semantic segmentation \cite{Mottaghi_CVPR14}, so researchers tried to improve the standard FCN by modeling or providing context in the network.
Liu \emph{et al.} \cite{liu2015parsenet} added global context features to a feature map by global pooling.
Yu \emph{et al.} \cite{YuKoltun2016} proposed dilation convolutions to aggregate wider context information.
In addition, graphical models are applied to model the relationship of neuron activation \cite{chen2014semantic, crfasrnn_iccv2015, liu2015semantic, lin2015efficient}.
Particularly, Chen \emph{et al.} \cite{chen2014semantic}  combined the strengths of conditional random field (CRF) with CNN to refine the prediction, and thus achieved more accurate results.
Zheng \emph{et al.} \cite{crfasrnn_iccv2015} formulated CRFs as recurrent neural networks (RNN), and trained the FCN and the CRF-RNN end-to-end.
Recurrent neural networks have also been used to replace graphical models in learning context dependencies \cite{byeon2015scene,shuai2016dag,li2016lstm}, which shows benefits in complicated scenarios. 

Recently, incorporating superpixels in convolutional networks has received much attention.
Superpixels are able to not only provide precise boundaries, but also to provide adaptive receptive fields.
For example, Dai \emph{et al.} \cite{dai2015convolutional} designed a convolutional feature masking layer for semantic segmentation, which allows networks to extract
features in unstructured regions with the help of superpixels.
Gadde \emph{et al.} \cite{raghudeep2015spCNN} improved the semantic segmentation using superpixel convolutional networks with bilateral inception, which can merge initial superpixels by parameters and generate different levels of regions.
Caesar \emph{et al.} \cite{region_end2end2016eccv} proposed a novel network with free-form ROI pooling which leverages superpixels to generate adaptive pooling regions.
Arnab \emph{et al.} \cite{arnab2016higher} modeled a CRF with superpixels as higher order potentials, and achieved better results than previous CRF based methods \cite{chen2014semantic, crfasrnn_iccv2015}.
Both methods showed the merit of providing superpixels to networks, which can generate more accurate segmentations.
Different from prior works \cite{raghudeep2015spCNN,region_end2end2016eccv}, we introduce superpixels at the end of convolutional networks instead of in the intermediate layers and also integrate the response from multiple views with average pooling, which has been used to replace the fully connected layers in classification \cite{nerworkinnetwork} and localization \cite{zhou2015cnnlocalization} tasks successfully.

\subsection{Semantic segmentation with videos}
The aim of multi-view semantic segmentation is to employ the potentially richer information from diverse views to improve over segmentations from a single view.
Couprie \emph{et al.} \cite{couprie2013indoor} performed single image semantic segmentation with learned features with color and depth information, and applied a temporal smoothing in test time to improve the performance of frame-by-frame estimations.
Hermans \emph{et al.} \cite{hermans2014dense} used the Bayesian update strategy to fuse new classification results and a CRF model in 3D space to smooth the segmentation.
{St{\"u}ckler \emph{et al.}  \cite{stuckler2015dense} used random forests to predict single view segmentations, and fused all views to the final output by a simultaneous localization and mapping (SLAM) system.}
Kundu \emph{et al.} \cite{kundu2016feature} built a dense 3D CRF model with correspondences from optical flow to refine semantic segmentation from video.
Recently, McCormac \emph{et al.} \cite{SemanticFusion} proposed a CNN based semantic 3D mapping system for indoor scenes. They applied a SLAM system to build correspondences, and mapped semantic labels predicted from CNN to 3D point cloud data.
Mustikovela \emph{et al.} \cite{Siva2016LabelProp} proposed to generate pseudo ground truth annotations for auxiliary data with a CRF based framework.
With the auxiliary data and their generated annotations, they achieved a clear improvement.
In contrast to the above methods, instead of integrating multi-view information by using graphical models, we utilize optical flow and image superpixels to establish region correspondences, and design a superpixel based multi-view network for semantic segmentation.

\section{Fully convolutional multi-view segmentation with region correspondences}
\label{sec:method}
\begin{figure}[!t]
\begin{center}
   \includegraphics[width=0.99\linewidth]{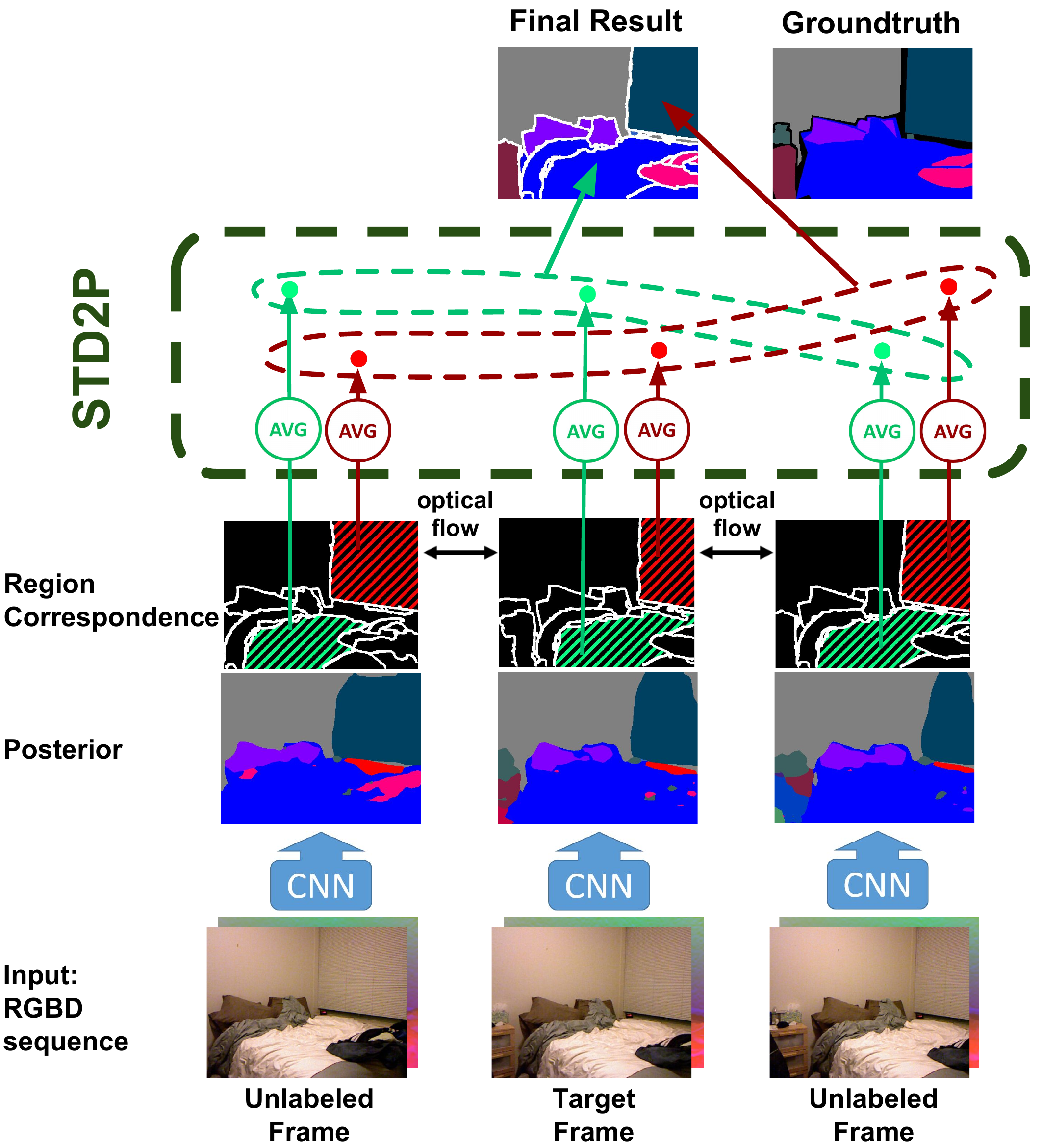}
\end{center}
   \caption{Pipeline of the proposed method. 
   Our multi-view semantic segmentation network is built on top of a CNN. It takes a RGBD sequence as input and computes the semantic segmentation of a target frame with the help of unlabeled frames.
   We use superpixels and optical flow to establish region correspondences, and fuse the posterior from multiple views with the proposed Spatio-Temporal Data-Driven Pooling (STD2P).
   }
\label{fig:pipeline}
\end{figure}

Our goal is a multi-view semantic segmentation scheme, that integrates seamlessly into exciting deep architectures and produces highly accurate semantic segmentation of a single view. We further aim at facilitating training from partially annotated input sequences, so that existing datasets can be used and the annotation effort stays moderate for new datasets. To this end, we draw on prior work on high quality non-semantic image segmentation and optical flow which is input to our proposed Spatio-Temporal Data-Driven Pooling (STD2P) layer.

\paragraph{Overview.} As illustrated in Figure~\ref{fig:pipeline}, our method starts from an image sequence. We are
interested in providing an accurate semantic segmentation of one view in the sequence, called {\it target frame}, 
which can be located at any position in the image sequence.
The two components that distinguish our approach from a standard fully convolutional architecture for semantic segmentation are, first, the computation of region correspondences and, second, the novel spatio-temporal pooling layer that is based on these correspondences.

We first compute the superpixel segmentation of each frame and establish region correspondences using optical flow. Then, the proposed data-driven pooling allows to aggregate information first within superpixels and then along their correspondences inside a CNN architecture. 
Thus, we achieve a tight integration of the superpixel segmentation and multi-view aggregation into a deep learning framework for semantic segmentation.

\subsection{Region correspondences}
\label{subsection:corr}
Motivated by the recent success of superpixel based approaches in deep learning architectures \cite{raghudeep2015spCNN,region_end2end2016eccv,arbelaez2012semantic,deng2015semantic} and the reduced computational load, we decide for a region-based approach. In the following, we motivate and detail our approach on establishing robust correspondences.

\paragraph{Motivation.}
One key idea of our approach is to map information from potentially unlabeled frames to the target frame, as diverse view points can provide additional context and resolve challenges in appearance and occlusion as illustrated in Figure \ref{fig:motivation}.
Hence, we do not want to assume visibility or correspondence of objects across all frames (e.g. the nightstand in the target frame as shown in Figure \ref{fig:pipeline}). Therefore, video supervoxel methods such as \cite{GrundmannKwatra2010} that force interframe correspondences and do not offer any confidence measure are not suitable. Instead, we establish the required correspondences on a frame-wise region level.

\paragraph{Superpixels \& optical flow.}
We compute RGBD superpixels \cite{gupta2014learning} in each frame to partition a RGBD image into regions,
and apply Epic flow \cite{revaud2015epicflow} between each pair of consecutive frames to link these regions. To take advantage of the depth information, we utilize the RGBD version of the structured edge detection \cite{dollar2013structured} to generate boundary estimates. Then, Epic flow is computed in forward and backward directions.

\begin{figure}[!t]
\begin{center}
   \includegraphics[trim=0cm 0.3cm 4.0cm 2.8cm, clip=true,width=0.45\linewidth]{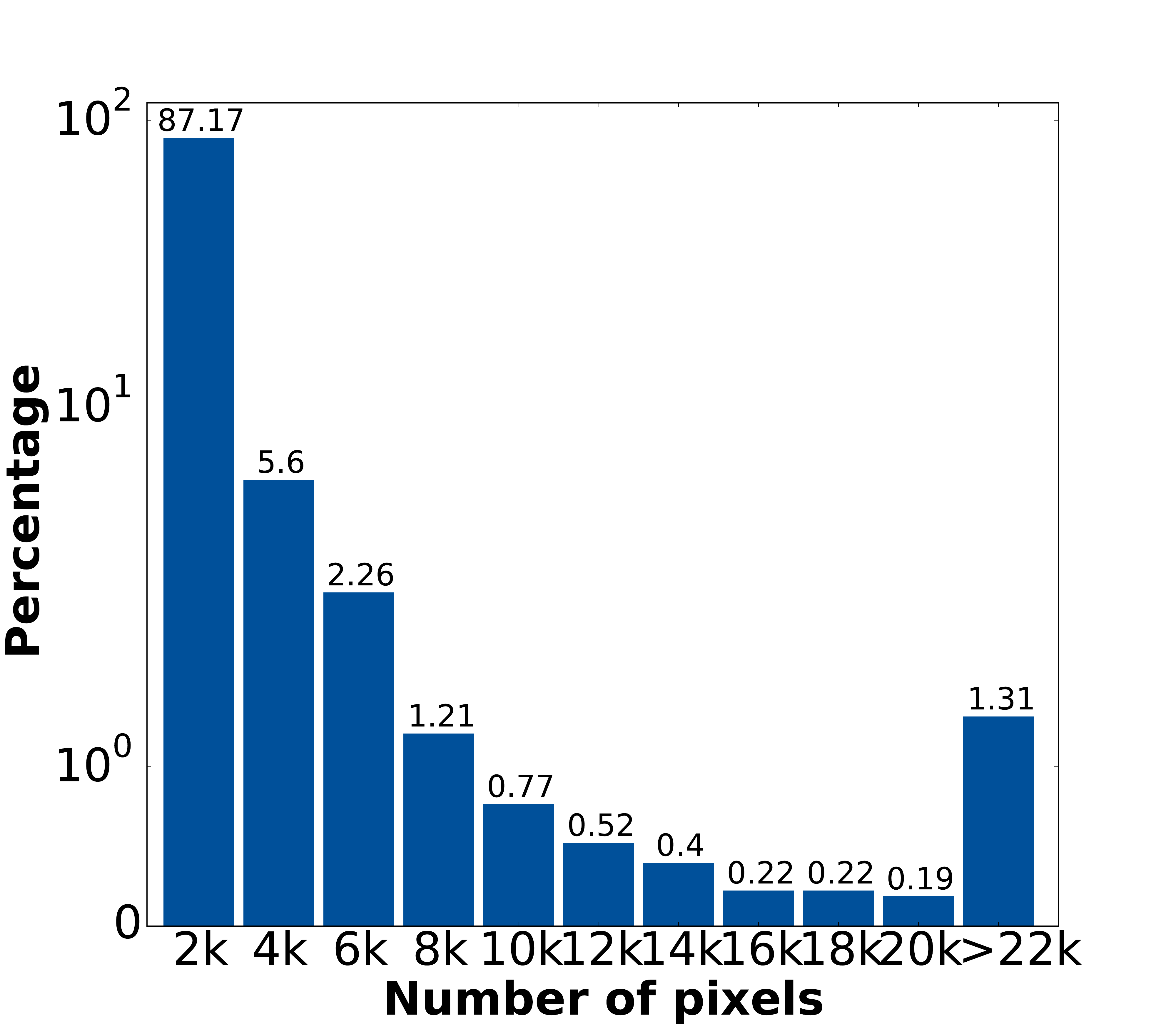}
   \includegraphics[trim=0.3cm 0.3cm 4.0cm 2.8cm, clip=true,width=0.45\linewidth]{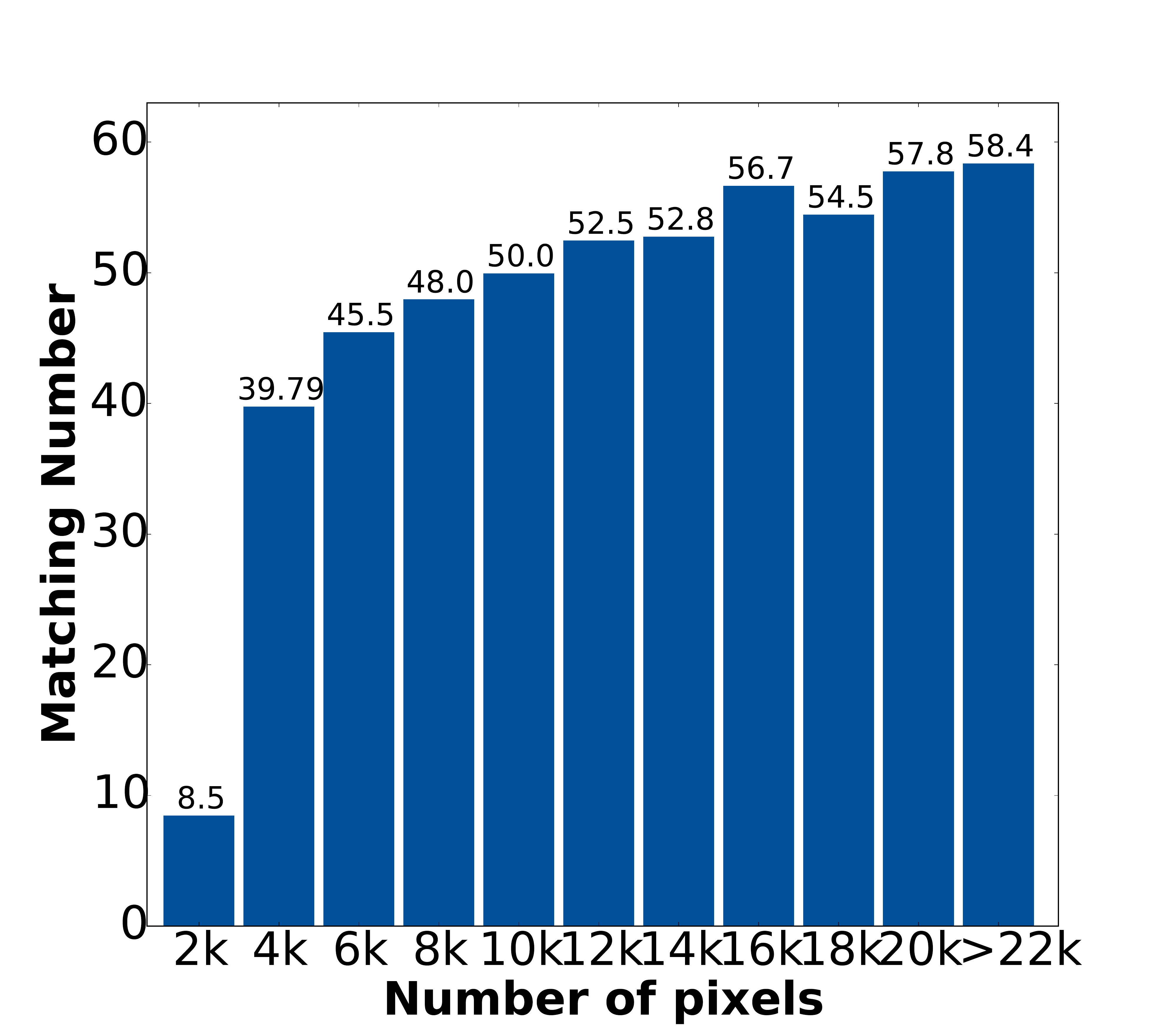}
   \end{center}
   \caption{Statistics of region correspondences on the NYUDv2 dataset. (left) Distribution of region sizes;
  (right) Histogram of the average number of matches over region sizes.}
\label{fig:statistics_region}
\end{figure}

\paragraph{Robust spatio-temporal matching.}
Given the precomputed regions in the target frame and all unlabeled frames as well as the optical flow between those frames, our goal is to find highly reliable region correspondences. For any two regions $R_t$ in the target frame $f_t$ and $R_u$ in an unlabeled frame $f_u$, we compute their matching score from their intersection over union (IoU). Let us assume w.l.o.g. that $u<t$. Then, we warp $R_u$ from $f_u$ to $R_u^{'}$ in $f_t$ using forward optical flow. The IoU between $R_t$ and $R_u^{'}$ is denoted by $\overrightarrow{IoU}_{tu}$. Similarly, we compute $\overleftarrow{IoU}_{tu}$ with backward optical flow.
We regard $R_t$ and $R_u$ as a successful match if their matching score meets $\min(\overleftarrow{IoU}_{tu},\overrightarrow{IoU}_{tu}) > \tau$. We keep the one with the highest matching score if $R_t$ has several successful matches. We show the statistics of region correspondences on the NYUDv2 dataset in Figure~\ref{fig:statistics_region}.

It shows that 87.17\% of the regions are relatively small (less than 2000 pixels) 
The plot on the right shows that those small regions generally only find less than 10 matches in a whole video.
Contrariwise, even slightly bigger regions can be matched more easily
 and they cover large portions of images. They usually have more than 40 matches in a whole video, and thus provide adequate information for our multi-view network.

\begin{figure*}[!t]
\begin{center}
   \includegraphics[trim=0cm 0cm 0cm 0cm, clip=true,width=0.99\linewidth]{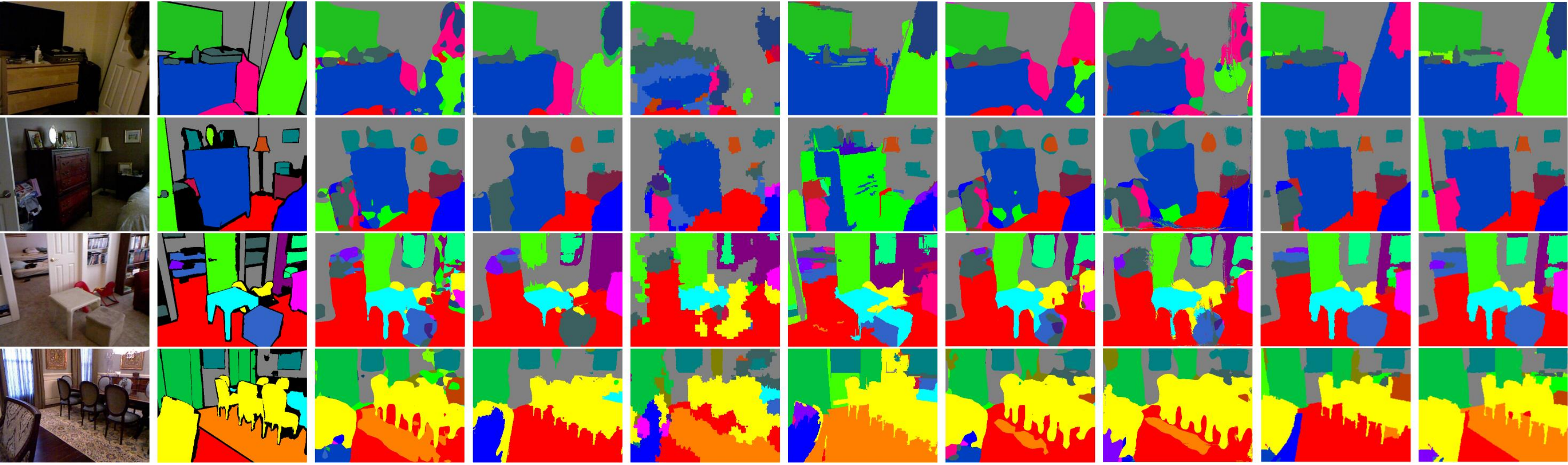}
\end{center}
   \scriptsize
   \hspace{0.7cm}Image
   \hspace{1.1cm}GT
   \hspace{0.98cm}CRF-RNN
   \hspace{0.45cm}DeepLab-LFOV
   \hspace{0.5cm}BI(3000)
   \hspace{1.0cm}E2S2
   \hspace{1.25cm}FCN
   \hspace{0.77cm}Singleview SP
   \hspace{0.18cm}Multiview Pixel
   \hspace{0.16cm}Our full model
   \caption{Visualization examples of the semantic segmentation on NYUDv2. Column 1 shows the RGB images and column 2
   shows the ground truth (black represents the unlabeled pixels).
   Columns 3 to 6 show the results from CRF-RNN \cite{crfasrnn_iccv2015}, DeepLab-LFOV \cite{chen2016deeplab}, BI(3000) \cite{raghudeep2015spCNN} and E2S2 \cite{region_end2end2016eccv}, respectively.
   Columns 7 to 9 show the results from FCN \cite{long2015fully}, single-view superpixel and multi-view pixel baselines. The results from our whole system are shown in column 10. Best viewed in color.}
\label{fig:baseline}
\end{figure*}

\subsection{Spatio-Temporal Data-Driven Pooling (STD2P)}
\label{subsec:network}
Here, we describe our Spatio-Temporal Data-Driven Pooling (STD2P) model that uses the spatio-temporal structure of the computed region correspondences to aggregate information across views as illustrated in Figure~\ref{fig:pipeline}.
While the proposed method is highly compatible with recent CNN and FCN models, we build on a per frame model using \cite{long2015fully}. 
In more detail, we refine the output of the deconvolution layer with superpixels and aggregate the information from multiple views by three layers: a spatial pooling layer, a temporal pooling layer and a region-to-pixel layer.

\paragraph{Spatial pooling layer.} The input to the spatial pooling layer is a feature map $I_s\in R^{N\times C\times H\times W}$ for $N$ frames,
$C$ channels with size $H\times W$ 
and a superpixel map $S\in R^{N\times H\times W}$ encoded with the region index.
It generates the output $O_s\in R^{N\times C\times P}$, where $P$ is the maximum number of superpixels.
The superpixel map $S$ guides the forward and backward propagation of the layer.
Here, $\Omega_{ij}=\{(x,y)|S(i,x,y)=j\}$ denotes a superpixel in the $i$-th frame with region index $j$. Then,
the forward propagation of spatial average pooling can be formulated as

\begin{equation}
\begin{split}
O_s(i,c,j) & = \frac{1}{|\Omega_{ij}|}\sum\limits_{(x,y)\in \Omega_{ij}}I_s(i,c,x,y)
\end{split}
\label{eq:forward_spatial}
\end{equation}
for each channel index $c$ of the $i$-th frame and region index $j$. 
We train our model using stochastic gradient descent.
The gradient of the input $I_s(i,c,x,y)$, where $(x,y)\in \Omega_{ij}$, 
in our spatial pooling is calculated by back propagation~\cite{rumelhart1988learning},

\begin{equation}
\begin{split}
\frac{\partial L}{\partial I_s(i,c,x,y)}  &=   \frac{\partial L}{\partial O_s(i,c,j)}\frac{\partial O_s(i,c,j)}{\partial I_s(i,c,x,y)} \\
&= \frac{1}{|\Omega_{ij}|} \frac{\partial L}{\partial O_s(i,c,j)}.
\end{split}
\label{eq:backward_spatial}
\end{equation}

\paragraph{Temporal pooling layer.} Similarly, we formulate our temporal pooling which fuses the information from $N$ frames $I_t\in R^{N\times C\times P}$, which is the output of spatial pooling layer, to one frame $O_t\in R^{C\times P}$. This layer also needs superpixel information $\Omega_{ij}$, which is the superpixel with index $j$ of the $i$-th input frame. If $\Omega_{ij}\neq\varnothing$, there exists correspondence.
The forward propagation can be expressed as

\begin{equation}
\begin{split}
O_t(c,j) & = \frac{1}{K}\sum\limits_{\Omega_{ij}\neq\varnothing}I_t(i,c,j)
\end{split}
\label{eq:forward_temporal}
\end{equation}
for channel index $c$ and region index $j$, where $K=|\{i |\Omega_{ij}\neq\varnothing, 1\le i\le N\}|$, which is the number of matched frames for $j$-th region.
The gradient is calculated by
\begin{equation}
\begin{split}
\frac{\partial L}{\partial I_t(i,c,j)}  &=  \frac{\partial L}{\partial O_t(c,j)}\frac{\partial O_t(c,j)}{\partial I_t(i,c,j)} \\
&= \frac{1}{K} \frac{\partial L}{\partial O_t(c,j)}.
\end{split}
\label{eq:backward_tempral}
\end{equation}

\paragraph{Region-to-pixel layer.} To directly optimize a semantic segmentation model with dense annotations,
we map the region based feature map $I_r\in R^{C\times P}$ to a dense pixel-level prediction $O_r\in R^{C\times H \times W}$.
This layer needs a superpixel map on the target frame $S_{\text{target}}\in R^{H\times W}$ to perform forward and backward propagation.
The forward propagation is expressed as
\begin{equation}
\begin{split}
O_r(c,x,y) & = I_r(c,j),\ \ \ S_{\text{target}}(x,y)=j.
\end{split}
\label{eq:forward_mapping}
\end{equation}
The gradient is computed by 
\begin{equation}
\begin{split}
\frac{\partial L}{\partial I_r(c,j)}  &=  \sum_{S_{\text{target}}(x,y)=j}\frac{\partial L}{\partial O_r(c,x,y)}\frac{\partial O_r(c,x,y)}{\partial I_r(c,j)} \\
&= \sum_{S_{\text{target}}(x,y)=j}\frac{\partial L}{\partial O_r(c,x,y)}.
\end{split}
\label{eq:backward_mapping}
 \end{equation}
   
\vspace{1.5cm}
\paragraph{Implementation details.}
\label{subsection:implementation}
We regard the frames with groundtruth annotations as target frames. For each target frame, we equidistantly sample up to 100 frames around it with the static interval of 3 frames.
Next, we compute the superpixels \cite{gupta2014learning} and Epic flow \cite{revaud2015epicflow} with the default settings provided in the corresponding source codes. The threshold $\tau$ for the computation of region correspondence is $0.4$ (cf. section~\ref{subsection:corr}).
Finally, for each RGBD sequence, we randomly sample 11 frames including the target frame together with their correspondence maps as the input for our network.
We use RGB images and HHA representations of depth \cite{gupta2014learning} and train the network by stochastic gradient descent with momentum term.
Due to the memory limitation, we first run FCN and cache the output \textit{pool4\_rgb} and \textit{pool4\_hha}. Then, we finetune the layers after \textit{pool4} with a new network which is the copy of the higher layers in FCN.
We use a minibatch size of 10, momentum 0.9, weight decay 0.0005 and fixed learning rate 10$^{-14}$.
We finetune our model by using cross entropy loss with 1000 iterations for all our models in the experiments.
{We implement the proposed network using the \textit{Caffe} framework \cite{jia2014caffe}, and the source code is available at \url{https://github.com/SSAW14/STD2P}}.

\section{Experiments and analysis}
\label{sec:experiments}

We evaluate our approach on the 4-class \cite{Silberman:ECCV12}, 13-class \cite{couprie2013indoor}, and 40-class \cite{gupta2013perceptual} tasks of the \textit{NYU--Depth--V2} (NYUDv2) dataset \cite{Silberman:ECCV12}, and 33-class task of the SUN3D dataset \cite{SUN3D}.

The NYUDv2 dataset contains 518 RGBD videos, which have more than 400,000 images.
Among them, there are 1449 densely labeled frames, which are split into 795 training images and 654 testing images.
{We follow the experimental settings of \cite{deng2015semantic} to test on 65 labeled frames.}
We compare our models of different settings to previous state-of-the-art multi-view methods as well as single-view methods, which are summarized in Table \ref{table:table_competing}.
We report the results on the labeled frames, using the same evaluation protocol and metrics as \cite{long2015fully},
pixel accuracy (\textit{Pixel Acc.}), mean accuracy (\textit{Mean Acc.}), region intersection over union (\textit{Mean IoU}),
and frequency weighted intersection over union (\textit{f.w. IoU}).

\begin{table}[h!]
\scriptsize
  \begin{center}
    \caption{Configurations of competing methods}
    \label{table:table_competing}
    \begin{tabular}{lcc}
      \toprule
       & RGB  & RGBD \\
      \cmidrule(lr){1-1}\cmidrule(lr){2-2}\cmidrule(lr){3-3}
      Single-View & \cite{david2015multiscale,alex2015bayesiansegnet}  & \cite{region_end2end2016eccv,chen2014semantic,chen2016deeplab,deng2015semantic,raghudeep2015spCNN,gupta2014learning,long2015fully,Unsupervised_RGBD_segmentation,specificfeature2016eccv,crfasrnn_iccv2015}   \\
      Multi-View  & / & \cite{couprie2013indoor,hermans2014dense,stuckler2015dense,SemanticFusion}  \\
      \bottomrule
    \end{tabular}
  \end{center}
\end{table}

\begin{table*}[t!]
\scriptsize
  \begin{center}
    \caption{Performance of the 40-class semantic segmentation task on NYUDv2. We compare our method to various state-of-the-art methods:
            \cite{long2015fully,gupta2014learning,alex2015bayesiansegnet,david2015multiscale} are also based on convolutional networks,
            \cite{chen2014semantic, crfasrnn_iccv2015, chen2016deeplab} are the models based on convolutional networks and CRF,
            and \cite{raghudeep2015spCNN, region_end2end2016eccv, deng2015semantic} are region labeling methods, and thus related to ours.
            We mark the best performance in all methods with \textbf{BOLD} font, and the second best one is written with \underline{UNDERLINE}.}
    \label{table:table_state_of_the_art}
    \begin{tabular}{lccccccccccccccc}
      \toprule
       Methods & \rotatebox{90}{wall}  & \rotatebox{90}{floor} & \rotatebox{90}{cabinet} & \rotatebox{90}{bed}  & \rotatebox{90}{chair} & \rotatebox{90}{sofa} & \rotatebox{90}{table}  & \rotatebox{90}{door} & \rotatebox{90}{window} & \rotatebox{90}{bookshelf}  & \rotatebox{90}{picture} & \rotatebox{90}{counter} & \rotatebox{90}{blinds}  & \rotatebox{90}{desk} & \rotatebox{90}{shelves} \\
      \cmidrule(lr){1-1}\cmidrule(lr){2-16}
      Mutex Constraints \cite{deng2015semantic}      & 65.6  & 79.2  & 51.9  & 66.7  & 41.0  & 55.7  & 36.5  & 20.3  & 33.2  & 32.6  & 44.6  & 53.6  & {49.1}  & 10.8  & \underline{9.1}  \\
      RGBD R-CNN  \cite{gupta2014learning}      & 68.0  & 81.3  & 44.9  & 65.0  & 47.9  & 47.9  & 29.9  & 20.3  & 32.6  & 18.1  & 40.3  & 51.3  & 42.0  & 11.3  & 3.5  \\
      Bayesian SegNet  \cite{alex2015bayesiansegnet}      & -  & -  & -  & -  & -  & -  & -  & -  & -  & -  & -  & -  & -  & -  & -  \\
      Multi-Scale CNN  \cite{david2015multiscale}      & -  & -  & -  & -  & -  & -  & -  & -  & -  & -  & -  & -  & -  & -  & -  \\
      CRF-RNN \cite{crfasrnn_iccv2015}      &  70.3  &  81.5  & 49.6  &  64.6  &  51.4  & 50.6  & 35.9   & 24.6   &  38.1  & 36.0  & 48.8   & 52.6   & 47.6   &  13.2  & 7.6   \\
      DeepLab \cite{chen2014semantic}      & 67.9  & 83.0  & 53.1  & 66.8  & 57.8  & 57.8  & \underline{43.4}  & 19.4  & \underline{45.5}  & 41.5  & 49.3  & \underline{58.3}  & 47.8  & 15.5  & 7.3  \\
      DeepLab-LFOV \cite{chen2016deeplab}      & 70.2  & \underline{85.2}  & \underline{55.3}  & 68.9  & \textbf{60.5}  & \underline{59.8}  & \textbf{44.5}  & 25.4  & \textbf{47.8}  & \textbf{42.6}  & 47.9  & 57.7  & \textbf{52.4}  & \textbf{20.7}  & \underline{9.1}  \\
      BI (1000)  \cite{raghudeep2015spCNN}      &  62.8  &  66.8  &  44.2  &  47.7  &  35.8  &  35.9  &  10.9  & 18.3   &  21.5  &  35.9  & 41.5   & 30.9   & 47.4   & 12.8 & 8.5  \\
      BI (3000) \cite{raghudeep2015spCNN}      &  61.7  &  68.1  &  45.2  &  50.6  &  38.9  &  40.3  &  26.2  & 20.9   &  36.0  &  34.4  & 40.8   & 31.6   & 48.3   & 9.3 & 7.9  \\
      E2S2  \cite{region_end2end2016eccv}  &  56.9  & 67.8   & 50.0  & 59.5  &  43.8  &   44.3  &  31.3  & 24.6   &  37.9  &  32.7  &  46.1 & 45.0  & \underline{51.8}   &  \underline{15.8}  &  \underline{9.1} \\
      FCN \cite{long2015fully}  & 69.9  & 79.4  & 50.3  & 66.0  & 47.5  & 53.2  & 32.8  & 22.1  & 39.0  & 36.1  & 50.5  & 54.2  & 45.8  & 11.9  & 8.6  \\
      \cmidrule(lr){1-1}\cmidrule(lr){2-16}
      Ours (\textit{superpixel}) & 70.9  & 83.4  & 52.6  & 68.5  & 54.1  & 56.0  & 40.4  & 25.5  & 38.4  & 40.9  & 51.5  & 54.8  & 47.3  & 11.3  & 7.5  \\
      Ours (\textit{superpixel+}) & \underline{72.4}  & {84.3}  & {52.0}  & \underline{71.5}  & {54.3}  & {58.8}  & {37.9}  & \underline{28.2}  & {41.9}  & {38.5}  & \underline{52.3}  & {58.2}  & {49.7}  & {14.3}  & 8.1  \\
      Ours (\textit{full model}) & \textbf{72.7}  & \textbf{85.7}  & \textbf{55.4}  & \textbf{73.6}  & \underline{58.5}  & \textbf{60.1}  & {42.7}  & \textbf{30.2}  & {42.1}  & \underline{41.9}  & \textbf{52.9}  & \textbf{59.7}  & 46.7  & {13.5}  & \textbf{9.4}  \\
     \hline
    \end{tabular}
    
    \begin{tabular}{lccccccccccccccc}
      Methods  & \rotatebox{90}{curtain}  & \rotatebox{90}{dresser} & \rotatebox{90}{pillow} & \rotatebox{90}{mirror}  & \rotatebox{90}{floormat} & \rotatebox{90}{clothes} & \rotatebox{90}{ceiling}  & \rotatebox{90}{books} & \rotatebox{90}{fridge} & \rotatebox{90}{tv}  & \rotatebox{90}{paper} & \rotatebox{90}{towel} & \rotatebox{90}{showercurtain}  & \rotatebox{90}{box} & \rotatebox{90}{whiteboard} \\
      \cmidrule(lr){1-1}\cmidrule(lr){2-16}
      Mutex Constraints \cite{deng2015semantic}      & \textbf{47.6}  & 27.6  & \textbf{42.5}  & {30.2}  & \underline{32.7}  & 12.6  & {56.7}  & 8.9  & 21.6  & 19.2  & \textbf{28.0}  & 28.6  & {22.9}  & 1.6  & 1.0  \\
      RGBD R-CNN \cite{gupta2014learning}      & 29.1  & 34.8  & 34.4  & 16.4  & 28.0  & 4.7  & \underline{60.5}  & 6.4  & 14.5  & 31.0  & 14.3  & 16.3  & 4.2  & 2.1  & 14.2  \\
      Bayesian SegNet \cite{alex2015bayesiansegnet}      & -  & -  & -  & -  & -  & -  & -  & -  & -  & -  & -  & -  & -  & -  & -  \\
      Multi-Scale CNN \cite{david2015multiscale}      & -  & -  & -  & -  & -  & -  & -  & -  & -  & -  & -  & -  & -  & -  & -  \\
      CRF-RNN \cite{crfasrnn_iccv2015}      &  34.8  &  33.2  & 34.7  & 20.8   &  24.0  & 18.7  &  \textbf{60.9}  & \underline{29.5}  & 31.2   & 41.1   & 18.2   & 25.6   &  \underline{23.0}  &  7.4  & 13.9   \\
      DeepLab \cite{chen2014semantic}      & 32.9  & 34.3  & 40.2  & 23.7  & 15.0  & \underline{20.2}  & 55.1  & 22.1  & 30.6  & {49.4}  & \underline{21.8}  & 32.1  & 6.4  & 5.8  & 14.8  \\ 
      DeepLab-LFOV \cite{chen2016deeplab}      & 36.0  & {36.9}  & 41.4  & \underline{32.5}  & 16.0  & 17.8  & 58.4  & 20.5  & \textbf{45.1}  & 48.0  & 21.0  & \textbf{41.5}  & 9.4  & \textbf{8.0}  & 14.3  \\ 
      BI (1000) \cite{raghudeep2015spCNN}      &  29.3  & 20.3  & 21.7   & 13.0   &  18.2  &  14.1  &  44.7  & 10.9   &  21.5  &  30.4  &  18.8  &  22.3  &  17.7  & 5.5 & 12.4   \\
      BI (3000) \cite{raghudeep2015spCNN}      &  30.8  & 22.9  & 19.5   & 13.9   &  16.1  &  13.7  &  42.5  & 21.3   &  16.6  &  30.9  &  14.9  &  23.3  &  17.8  & 3.3 & 9.9   \\
      E2S2 \cite{region_end2end2016eccv}     & 38.0 &  34.8  & 31.5  &  31.7  &  25.3  &   14.2 & 39.7 & 26.7   & 27.1   &  35.2  &  17.8  &  21.0 &  19.9  & 7.4   & \textbf{36.9}    \\
      FCN  \cite{long2015fully}  & 32.5  & 31.0  & 37.5  & 22.4  & 13.6  & 18.3  & {59.1}  & 27.3  & 27.0  & 41.9  & 15.9  & 26.1  & 14.1 & {6.5}  & 12.9  \\
      \cmidrule(lr){1-1}\cmidrule(lr){2-16}
      Ours (\textit{superpixel}) & 34.5  & \underline{41.6}  & 37.7  & 20.1  & 15.9  & 20.1  & 56.8  & 28.8  & 23.8  & \underline{51.8} & 19.1  & 26.6  & \textbf{29.3} & {6.8}  & {4.7}  \\
      Ours (\textit{superpixel+}) & \underline{42.9}  & {35.9}  & 40.8  & 27.7  & 31.9  & {19.3}  & 55.6  & {28.2}  & {38.3}  & {46.9}  & 17.6  & {31.2}  & 11.0 & {6.5}  & {28.2}  \\
      Ours (\textit{full model}) & 40.7  & \textbf{44.1}  & \underline{42.0}  & \textbf{34.5}  & \textbf{35.6}  & \textbf{22.2}  & 55.9  & \textbf{29.8}  & \underline{41.7}  & \textbf{52.5}  & {21.1}  & \underline{34.4}  & {15.5} & \underline{7.8}  & \underline{29.2}  \\
      \hline
    \end{tabular}

    \begin{tabular}{lcccccccccccccc}
       Methods & \rotatebox{90}{person}  & \rotatebox{90}{nightstand} & \rotatebox{90}{toilet} & \rotatebox{90}{sink}  & \rotatebox{90}{lamp} & \rotatebox{90}{bathtub} & \rotatebox{90}{bag}  & \rotatebox{90}{other struct} & \rotatebox{90}{other furni} & \rotatebox{90}{other props}  & \rotatebox{90}{Pixel Acc.} & \rotatebox{90}{Mean Acc.} & \rotatebox{90}{Mean IoU} & \rotatebox{90}{f.w. IoU}  \\
      \cmidrule(lr){1-1}\cmidrule(lr){2-11}\cmidrule(lr){12-15}
      Mutex Constraints \cite{deng2015semantic}      & 9.6 & 30.6  & 48.4  & 41.8  & 28.1  & 27.6  & 0  & 9.8  & 7.6  & 24.5  & 63.8  & - & 31.5  & 48.5  \\
      RGBD R-CNN \cite{gupta2014learning}     & 0.2  & 27.2  & 55.1  & 37.5  & 34.8  & {38.2}  & 0.2  & 7.1  & 6.1  & 23.1  & 60.3  & - & 28.6  & 47.0 \\
      Bayesian SegNet \cite{alex2015bayesiansegnet}      & -  & -  & -  & -  & -  & -  & -  & -  & -  & -  & 68.0  & 45.8 & 32.4  & - \\
      Multi-Scale CNN \cite{david2015multiscale}      & -  & -  & -  & -  & -  & -  & -  & -  & -  & -  & 65.6  & 45.1 & 34.1  & 51.4  \\
      CRF-RNN \cite{crfasrnn_iccv2015}      & 57.9 &  31.4  & 57.2  &  45.4  & 36.9  &  39.1  & 4.9  & 14.6   &  9.5  &  29.5  &  66.3  & 48.9  & 35.4   & 51.0   \\
      DeepLab \cite{chen2014semantic}      & 55.3 & 37.7  & 57.9  & \underline{47.7}  & \underline{40.0}  & \underline{44.7}  & 6.6  & 18.0  & \underline{12.9}  & \textbf{33.8}  & 68.7  & 46.9 & 36.8  & 52.5  \\
      DeepLab-LFOV \cite{chen2016deeplab}      & \textbf{67.0} & \underline{41.8}  & \textbf{69.7}  & 46.8  & \textbf{40.1}  & \textbf{45.1}  & 2.1  & \textbf{20.7}  & 12.4  & \underline{33.5}
      & \textbf{70.3}  & 49.6 & \underline{39.4}  & \underline{54.7}  \\
      BI (1000) \cite{raghudeep2015spCNN}        &  45.9  &  15.8 & 56.5   & 32.2  & 24.7   & 17.1  &  0.1  &  12.2  & 6.7 & 21.9   & 57.7   & 37.8  & 27.1   & 41.9   \\
      BI (3000) \cite{raghudeep2015spCNN}        &  44.7  &  15.8 & 53.8   & 32.1  & 22.8   & 19.0  &  0.1  &  12.3  & 5.3 & 23.2   & 58.9   & 39.3  & 27.7   & 43.0   \\
      E2S2 \cite{region_end2end2016eccv}      & 35.0 & 17.6 &  31.8  & 36.3  &  14.8  &  26.0 &  \textbf{9.9}  & 14.5 &   {9.3}  & 20.9   & 58.1   & \underline{52.9}  & 31.0   & 44.2   \\
      FCN  \cite{long2015fully}  & 57.6  & 30.1  & 61.3  & 44.8  & 32.1  & {39.2}  & 4.8  & 15.2  & 7.7  & {30.0}  & 65.4  & 46.1 & 34.0  & 49.5  \\
      \cmidrule(lr){1-1}\cmidrule(lr){2-11}\cmidrule(lr){12-15}
      Ours (\textit{superpixel}) & 66.1  & {37.4}  & 56.1  & 46.3  & 34.5  & 26.7  & 5.8  & 12.7  & 12.3  & 30.6  & {68.5}  & {48.7} & {36.0}  & {52.9} \\
      Ours (\textit{superpixel+}) & \underline{66.7}  & {34.1}  & \underline{62.8}  & \textbf{47.8}  & {35.1}  & 26.4  & \underline{8.8}  & \underline{19.3}  & {10.9}  & {29.2}  & {68.4}  & {52.1} & {38.1}  & {54.0} \\
      Ours (\textit{full model}) & {60.7}  & \textbf{42.2}  & {62.7}  & {47.4}  & {38.6}  & {28.5}  & {7.3}  & {18.8}  & \textbf{15.1}  & {31.4}  & \underline{70.1}  & \textbf{53.8} & \textbf{40.1}  & \textbf{55.7} \\
     \bottomrule
    \end{tabular}
  \end{center}
\end{table*}

\subsection{Results on NYUDv2 40-class task}
Table \ref{table:table_state_of_the_art} evaluates performance of our method on NYUDv2 40-class task
and compares to state-of-the-art methods and related approaches \cite{long2015fully,deng2015semantic,gupta2014learning,alex2015bayesiansegnet,david2015multiscale,crfasrnn_iccv2015,chen2014semantic,chen2016deeplab,raghudeep2015spCNN,region_end2end2016eccv}
\footnote{For \cite{long2015fully,deng2015semantic, gupta2014learning, alex2015bayesiansegnet, david2015multiscale}, we copy the performance from their paper. For \cite{crfasrnn_iccv2015,chen2014semantic,chen2016deeplab,raghudeep2015spCNN,region_end2end2016eccv}, we run the code provided by the authors with RGB+HHA images.
Specifically, for \cite{raghudeep2015spCNN}, we also increase the maximum number of superpixels from 1000 to 3000. The original coarse version and the fine version are abbreviated as BI(1000) and BI(3000).
}.
We include 3 versions of our approach:

\paragraph{Our \textit{superpixel} model} is trained on single frames without additional unlabeled data, and tested using a single target frame. It improves the baseline FCN on all four metrics by at least 2 percentage points (\textit{pp}),
and it achieves in particular better performance than recently proposed methods based on superpixels and CNN\cite{raghudeep2015spCNN,region_end2end2016eccv}.
\paragraph{Our \textit{superpixel+} model} leverages additional unlabeled data in the training while it only uses the target frame for test. It obtains 3.4\textit{pp}, 2.1\textit{pp}, 1.1\textit{pp} improvements over the \textit{superpixel} model on \textit{Mean Acc.}, \textit{Mean IoU} and \textit{f.w. IoU},
leading to more favorable performance than many state-of-the-art methods \cite{deng2015semantic,gupta2014learning,alex2015bayesiansegnet,david2015multiscale,crfasrnn_iccv2015,chen2014semantic,raghudeep2015spCNN,region_end2end2016eccv}.
This highlights the benefits of leveraging unlabeled data.
\paragraph{Our \textit{full model}} leverages additional unlabeled data both in the training and test. It achieves a consistent improvement over the \textit{superpixel+} model and outperforms all competitors in \textit{Mean Acc.}, \textit{Mean IoU} and \textit{f.w. IoU} by $0.9pp, 0.7pp, 1.0pp$ respectively.
Particularly strong improvements are observed on challenging object classes such as dresser(+$7.2pp$), door(+$4.8pp$), bed(+$4.7pp$) and TV(+$3.1pp$).

Figure \ref{fig:baseline} demonstrates that our method is able to produce smooth predictions with accurate boundaries.
We present the most related methods, which either apply CRF \cite{crfasrnn_iccv2015, chen2016deeplab} or incorporate superpixels \cite{raghudeep2015spCNN, region_end2end2016eccv}, in the columns 3 to 6 of this figure.
According to the qualitative comparison to these approaches, we can see the benefit of our method. It captures small objects like chair legs, as well as large areas like floormat and door.
In addition, we also present FCN and the \textit{superpixel} model at the 7-th and 8-th column of Figure~\ref{fig:baseline}. The FCN is boosted by introducing superpixels but not as precise as our \textit{full model} using unlabeled data.

\vspace{-0.3cm}
\paragraph{Average vs. max spatio-temporal data-driven pooling.}
Our data-driven pooling aggregates the local information from multiple observations within a segment and across multiple views.
Average pooling and max pooling are canonical choices used in many deep neural network architectures.
Here we test average pooling and max pooling both in the spatial and temporal pooling layer, and show the results in Table~\ref{table:table_avg_max}.
All the models are trained with multiple frames, and tested on multiple frames.
Average pooling turns out to perform best for spatial and temporal pooling. This result confirms our design choice.

\begin{table}[t!]
\scriptsize
  \begin{center}
    \caption{Comparison of average and max spatio-temporal data-driven pooling.}
    \label{table:table_avg_max}
    \begin{tabular}{cccccc}
      \toprule
      Spatial/Temporal & Pixel Acc.  & Mean Acc. & Mean IoU & f.w. IoU \\
      \cmidrule(lr){1-1}\cmidrule(lr){2-5}
      {\sc Avg} / {\sc Avg} & \textbf{70.1} & \textbf{53.8} & \textbf{40.1} & \textbf{55.7}  \\
      {\sc Avg} / {\sc Max}  & 69.4 & 51.0 & 38.0  & 54.4 \\
      {\sc Max} / {\sc Avg} & 66.4 & 45.4 & 33.8 & 49.6 \\
      {\sc Max} / {\sc Max} & 64.9 & 44.5 & 32.1 & 47.9 \\
      \bottomrule
    \end{tabular}
  \end{center}
\end{table}

\vspace{-0.2cm}
\paragraph{Region vs. pixel correspondences.}
We compare our \textit{full model}, which is built on the region correspondences, to the model with pixel correspondences. It only uses the per-pixel correspondences by optical flow and applies average pooling to fuse the information from multiple view. The visualization results of this baseline are presented in column 9 of Figure~\ref{fig:baseline}. 
Obtaining accurate pixel correspondences is challenging because the optical flow is not perfect and the error can accumulate over time.
Consequently, the model with pixel correspondences only improves slightly over the FCN baseline, as it is also reflected in the numbers in Table~\ref{table:table_baseline}.
Establishing region correspondences with the proposed rejection strategy described in section \ref{subsection:corr} seems indeed to be favorable over pixel correspondences.
Our \textit{full model} shows a significant improvement over the pixel-correspondence baseline and FCN in all 4 measures. 

\begin{figure}[!t]
\begin{center}
   \includegraphics[trim=0.1cm 0cm 3.5cm 2.5cm, clip=true,width=0.48\linewidth]{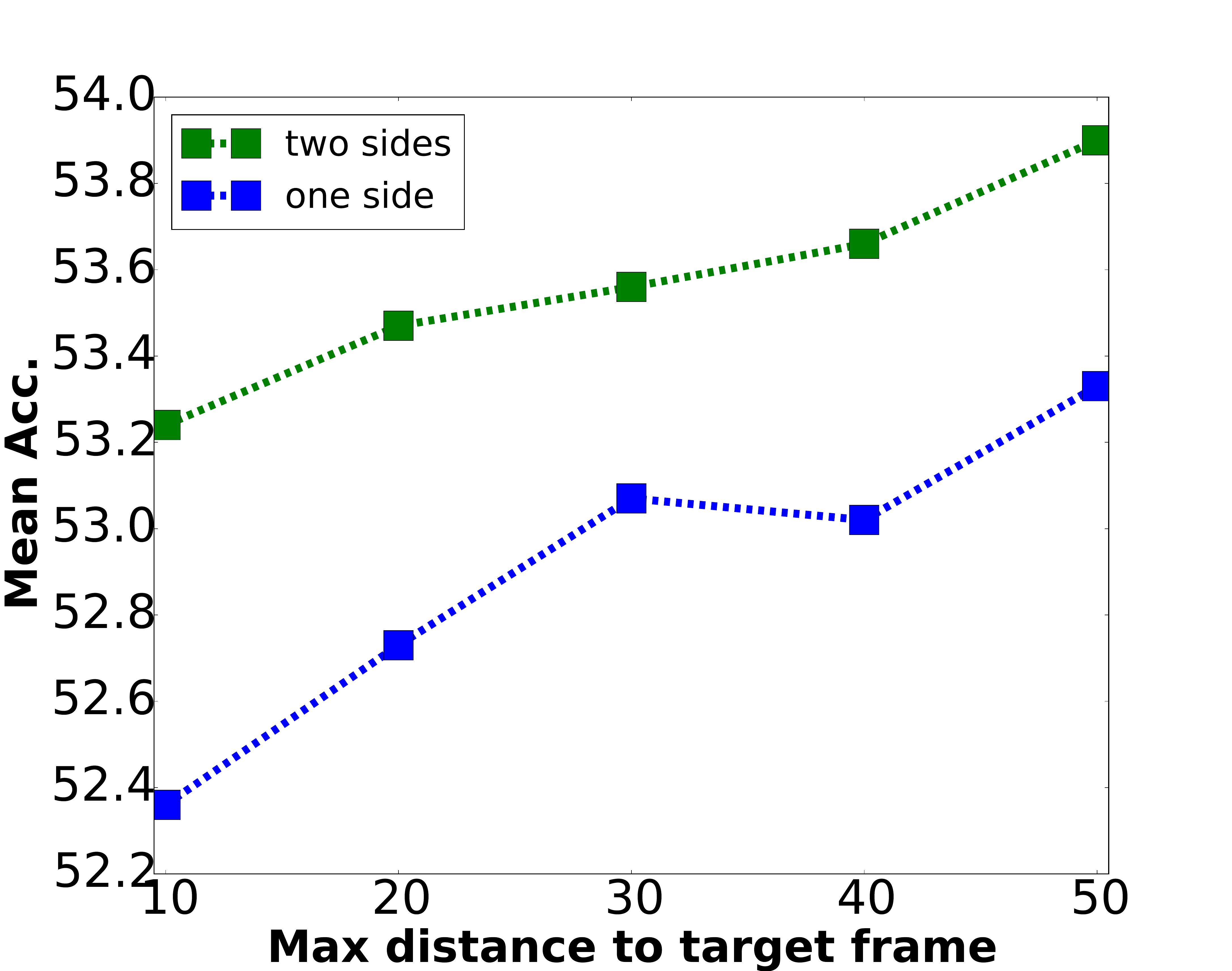}
   \includegraphics[trim=0.1cm 0cm 3.5cm 2.5cm, clip=true,width=0.48\linewidth]{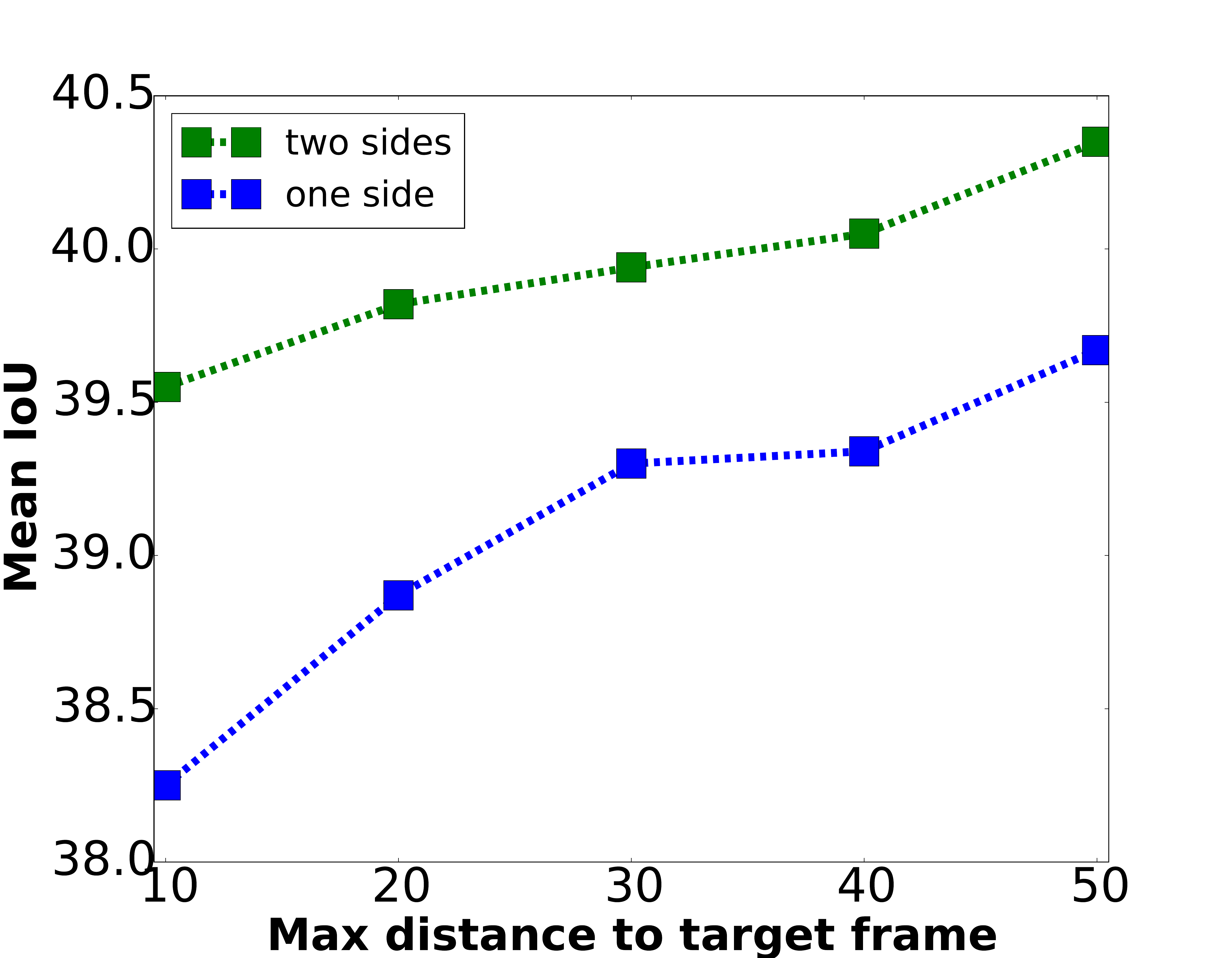}
   \end{center}
   \caption{The performance of multi-view prediction with varying maximum distance.
   Green lines show the results of using future and past views.
   Blue lines show the results of only using past views. }
\label{fig:statistics_max_k}
\end{figure}

\vspace{-0.3cm}
\paragraph{Analysis of multi-view prediction.}
In our multi-view model, we subsample frames from a whole video for computational considerations.
There is a trade-off between close-by and distant frames to be made.
If we select frames far away from the target frames, they can provide more diverse views of an object, while matching is more challenging and potentially less accurate than for close-by frames.
Hence, we analyze the influence of the distance of selected frames to target frames, and report the \textit{Mean Acc.} and \textit{Mean IoU} in Figure \ref{fig:statistics_max_k}.
In results, providing wider views is helpful, as the performance is improved with the increase of max distance.
And selecting the data in the future, which is another way to provide wider views, also contributes to the improvements of performance.

\begin{table}[t!]
\scriptsize
  \begin{center}
    \caption{Comparison results with baselines on NYUDv2 40-class task}
    \label{table:table_baseline}
    \begin{tabular}{lcccc}
      \toprule
      Methods & Pixel Acc.  & Mean Acc. & Mean IoU & f.w. IoU \\
      \cmidrule(lr){1-1}\cmidrule(lr){2-5}
      FCN \cite{long2015fully} & 65.4  & 46.1 & 34.0  & 49.5  \\
      Pixel Correspondence  & 66.2 & 45.9 & 34.6  & 50.2 \\
      Superpixel Correspondence & \textbf{70.1} & \textbf{53.8} & \textbf{40.1} & \textbf{55.7} \\
      \bottomrule
    \end{tabular}
  \end{center}
\end{table}

\begin{figure*}[!t]
\begin{center}
   \includegraphics[width=0.78\linewidth]{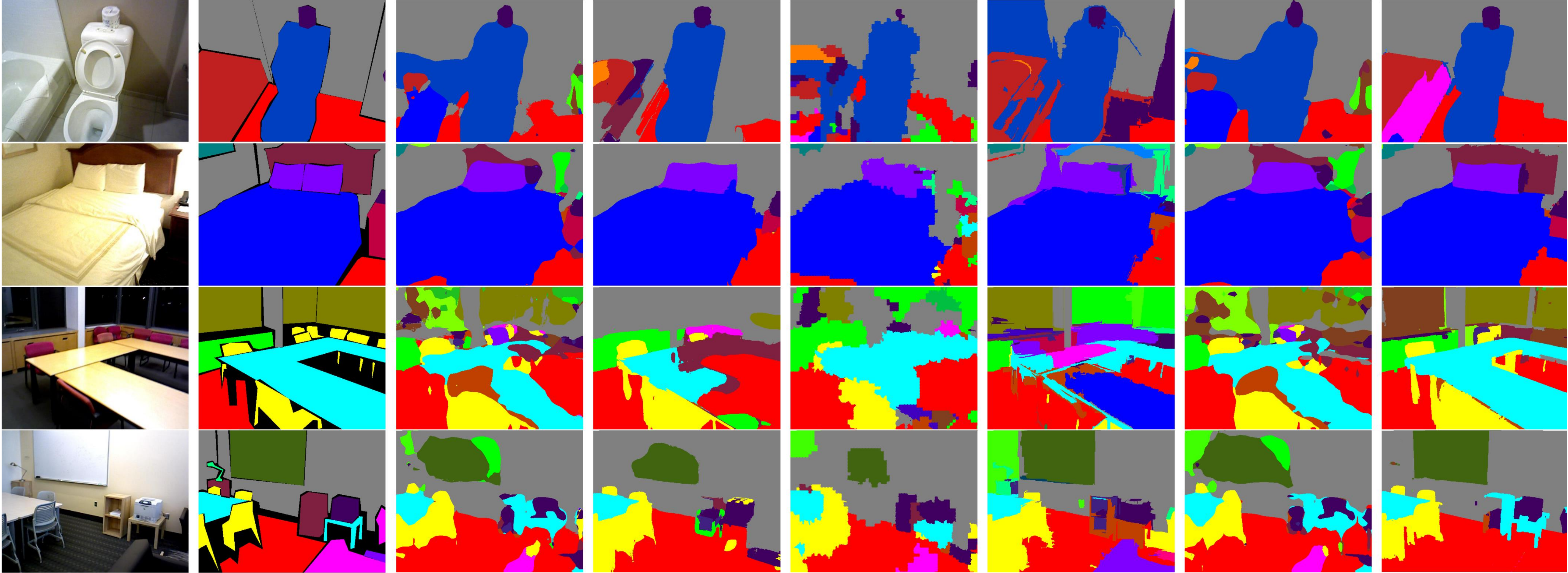}
   \end{center}
   \scriptsize
   \hspace{2.4cm} Image
   \hspace{1.1cm}GT
   \hspace{0.95cm}CRF-RNN
   \hspace{0.4cm}DeepLab-LFOV
   \hspace{0.48cm}BI(3000)
   \hspace{1.01cm}E2S2
   \hspace{1.16cm}FCN
   \hspace{0.77cm}Our full model
   \caption{Qualititive results of the SUN3D dataset.
   For each example, the images are arranged from top to bottom, from left to right as color image, groundtruth, CRF-RNN \cite{crfasrnn_iccv2015}, DeepLab-LFOV \cite{chen2016deeplab}, BI \cite{raghudeep2015spCNN}, E2S2 \cite{region_end2end2016eccv}, FCN \cite{long2015fully} and ours.}
\label{fig:sun3d}
\end{figure*}

\subsection{Results on NYUDv2 4-class and 13-class tasks}
\vspace{-0.1cm}
To show the effectiveness of our multi-view semantic segmentation approach, we compare our method to previous state-of-the-art multi-view semantic segmentation methods \cite{couprie2013indoor,hermans2014dense,stuckler2015dense,SemanticFusion} on the 4-class and 13-class tasks of NYUDv2 as shown in Table \ref{table:table_state_of_the_art_multiview}.
Besides, we also present previous state-of-the-art single-view methods  \cite{david2015multiscale, specificfeature2016eccv,Unsupervised_RGBD_segmentation}.
We observe that our \textit{superpixel+} model already outperforms all the multi-view competitors, and the proposed temporal pooling scheme further boosts \textit{Pixel Acc.} and \textit{Mean Acc.} by more than $1pp$ and then outperforms the state-of-the-art \cite{david2015multiscale}.
In particular, the recent proposed method by McCormac \emph{et al.} \cite{SemanticFusion} is also built on CNN, however, their performance on 13-class task is about $5pp$ worse than ours.

\begin{table}[t!]
\scriptsize
  \begin{center}
    \caption{Performance of the 4-class (left) and 13-class (right) semantic segmentation tasks on NYUDv2. }
    \label{table:table_state_of_the_art_multiview}
    \begin{tabular}{lcccc}
      \toprule
       Methods & {Pixel Acc.} & {Mean Acc.} & {Pixel Acc.} & {Mean Acc.} \\
      \cmidrule(lr){1-1}\cmidrule(lr){2-3}\cmidrule(lr){4-5}
      Couprie \emph{et al.} \cite{couprie2013indoor}  & 64.5  & 63.5  &  52.4  & 36.2 \\
      Hermans \emph{et al.} \cite{hermans2014dense}     & 69.0  & 68.1 & 54.2 & 48.0 \\
      St{\"u}ckler \emph{et al.} \cite{stuckler2015dense}     & 70.6  & 66.8  & - & - \\
      McCormac \emph{et al.} \cite{SemanticFusion}      & -  & -   & 69.9 & 63.6 \\
      \cmidrule(lr){1-1}\cmidrule(lr){2-3}\cmidrule(lr){4-5}
      Wang \emph{et al.} \cite{Unsupervised_RGBD_segmentation}      & -  & 65.3   & - & 42.2 \\
      Wang \emph{et al.} \cite{specificfeature2016eccv}      & -  & 74.7   & - & 52.7 \\
      Eigen \emph{et al.} \cite{david2015multiscale}      & \underline{83.2}  & \underline{82.0}   & \underline{75.4} & 66.9 \\
      \cmidrule(lr){1-1}\cmidrule(lr){2-3}\cmidrule(lr){4-5}
      Ours (\textit{superpixel+}) & {82.7}  & {81.3}   & 74.8 & \underline{67.0} \\
      Ours (\textit{full model}) & \textbf{83.6}  & \textbf{82.5}   & \textbf{75.8} & \textbf{68.4} \\
      \bottomrule
    \end{tabular}
  \end{center}
\end{table}

\vspace{-0.1cm}
\subsection{Results on SUN3D 33-class task}
\vspace{-0.15cm}
Table \ref{table:sun3d} shows the results of our method and baselines on the SUN3D dataset.
We follow the experimental settings of \cite{deng2015semantic} to test all the methods \cite{deng2015semantic,crfasrnn_iccv2015,chen2014semantic,chen2016deeplab,raghudeep2015spCNN,region_end2end2016eccv,long2015fully} on all 65 labeled frames in SUN3D,
which are trained with the NYUDv2 40-class annotations.
After computing the 40-class prediction, we map 7 unseen semantic classes into 33 classes.
Specifically, \textit{floormat} is merged to \textit{floor}, \textit{dresser} is merged to  \textit{other furni} and five other classes are merged to \textit{other  props}.
Among all the methods, we achieve the best \textit{Mean IoU} score that our \textit{superpixel+} and \textit{full model} are $1.2pp$ and $4.7pp$ better than \cite{deng2015semantic} and \cite{chen2016deeplab} .
For \textit{Pixel Acc.}, our method is comparable to the previous state of the art \cite{deng2015semantic}.
In addition, we observe that our \textit{superpixel+} model boosts the baseline FCN by $3.7pp$, $2.3pp$, $3.3pp$, $3.9pp$ on the four metrics, and applying multi-view information further improves $3.0pp$, $0.4pp$, $3.5pp$, $3.7pp$, respectively.
Besides, we achieve much better performance than DeepLab-LFOV, which is comparable to our model on the NYUDv2 40-class task.
This illustrates  the generalization capability of our model, even without finetuning on the new  domain or dataset.

\begin{table}[t!]
\scriptsize
  \begin{center}
    \caption{Performance of the 33-class semantic segmentation task on SUN3D. All 65 images are used as the test set. }
    \label{table:sun3d}
    \begin{tabular}{lcccc}
      \toprule
      Methods & Pixel Acc. & Mean Acc. & Mean IoU & f.w. IoU \\
      \cmidrule(lr){1-1}\cmidrule(lr){2-5}
      Mutex Constraints \cite{deng2015semantic}     & \textbf{65.7}  & - & 28.2 & \underline{51.0} \\
      CRF-RNN \cite{crfasrnn_iccv2015}     & 59.8  & -  & 25.5 & 43.3 \\
      DeepLab \cite{chen2014semantic}      & 60.9  & 30.7   & 24.0 & 44.1 \\
      DeepLab-LFOV \cite{chen2016deeplab}    & 62.3  & 35.3   & 28.2 & 46.2 \\
      BI (1000) \cite{raghudeep2015spCNN}   & 53.8 & 31.1 & 20.8 & 37.1 \\
      BI (3000) \cite{raghudeep2015spCNN}   & 53.9 & 31.6 & 21.1 & 37.4 \\
      E2S2 \cite{region_end2end2016eccv}     & 56.7  & \textbf{47.7}  & 27.2  & 43.3  \\
      FCN \cite{long2015fully}     & 58.8  & 38.5 & 26.1 & 43.9 \\
      \cmidrule(lr){1-1}\cmidrule(lr){2-5}
      Ours (\textit{superpixel+}) & {62.5}  & 40.8   & \underline{29.4} & {47.8} \\
      Ours (\textit{full model}) & \underline{65.5}  & \underline{41.2}   & \textbf{32.9} & \textbf{51.5} \\
      \bottomrule
    \end{tabular}
  \end{center}
\end{table}

\vspace{-0.2cm}
\section{Conclusion}
\vspace{-0.1cm}
\label{sec:conclusion}
We have presented a novel semantic segmentation approach using image sequences.
We design a superpixel-based multi-view semantic segmentation network with spatio-temporal data-driven pooling which can receive multiple images and their correspondence as input.
We propagate the information from multiple views to the target frame, and significantly improve the semantic segmentation performance on the target frame.
Besides, our method can leverage large scale unlabeled images for training and test, and we show that using unlabeled data also benefits single image semantic segmentation.

\vspace{-0.1cm}
\section*{Acknowledgments}
\vspace{-0.1cm}
This research was supported by the German Research Foundation  (DFG  CRC  1223) and
the ERC Starting Grant VideoLearn. 

{\small
\bibliographystyle{ieee}
\bibliography{egbib}
}

\clearpage

\section*{Supplementary Materials}
\renewcommand\thesubsection{\Alph{subsection}}
\subsection{Analysis of semantic segmentation boundary accuracy}
\begin{figure}[h!]
\begin{center}
   \includegraphics[trim=3.6cm 8.2cm 4cm 8.5cm, clip=true,width=0.9\linewidth]{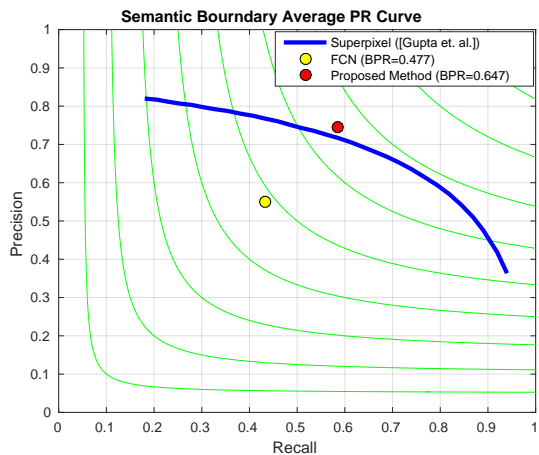}
\end{center}
   \caption{Precision-recall curve on semantic boundaries on the NYUDv2 dataset.}
\label{fig:boundary_bpr}
\end{figure}

In order to quantify the improvement on semantic boundary localization based on the proposed data-driven pooling scheme,
we use Boundary Precision Recall (BPR), as also used in image or video segmentation benchmark~\cite{GalassoetalICCV13,arbelaez2011contour} for evaluation.
Figure \ref{fig:boundary_bpr} shows the resulting semantic boundary average precision-recall curve. 
We conclude that our method generates  more accurate boundaries than FCN,
which achieve 0.477 BPR score while our method achieves 0.647.
Besides, our method even improves on the superpixel \cite{gupta2014learning} we build on,
which means our method can successfully merge over-segmentations or non-semantic boundaries between adjacent instances of the same semantic class.

\subsection{Oracle performance using groundtruth labels}
\begin{table}[h!]
\scriptsize
  \begin{center}
    \caption{The performance of oracle case using groundtruth to label the regions.}
    \label{table:table_oracle}
    \begin{tabular}{lcccc}
      \toprule
      Groundtruth & Pixel Acc.  & Mean Acc. & Mean IoU & f. w. IoU \\
      \cmidrule(lr){1-1}\cmidrule(lr){2-5}
      Current Frame & 96.2 & 94.0 & 90.2 &  92.7  \\
      Next Frame & 84.7 & 76.2 & 63.4 & 74.4   \\
      \bottomrule
    \end{tabular}
  \end{center}
\end{table}

We perform two best-case analysis by computing an oracle performance where groundtruth labels are available for either reference or target frames.
The first row of Table \ref{table:table_oracle} shows the achievable performance by performing a majority vote of the groundtruth pixel labels on the employed superpixels from \cite{gupta2014learning}. Thereby we achieve an upper bound of $96.2\%$ on the pixel accuracy that is implied by the superpixel over-segmentation.
In order to evaluate the effectiveness of our region correspondence, we use groundtruth labels of reference frames in the sequence.
We collect 143 views to conduct this experiment in NYUDv2, which have corresponding regions in target frames.
We ignore regions without correspondence in the next frame to compute the quantitative results, which are presented in Table \ref{table:table_oracle}.
This best-case analysis for correspondence results in a pixel accuracy of $84.7\%$.
Both oracle performances indicate a strong potential for performance improvements in our setup in all 4 reported measures.

\subsection{Groundtruth analysis}
\begin{figure*}[!t]
\begin{center}
   \includegraphics[width=0.8\linewidth]{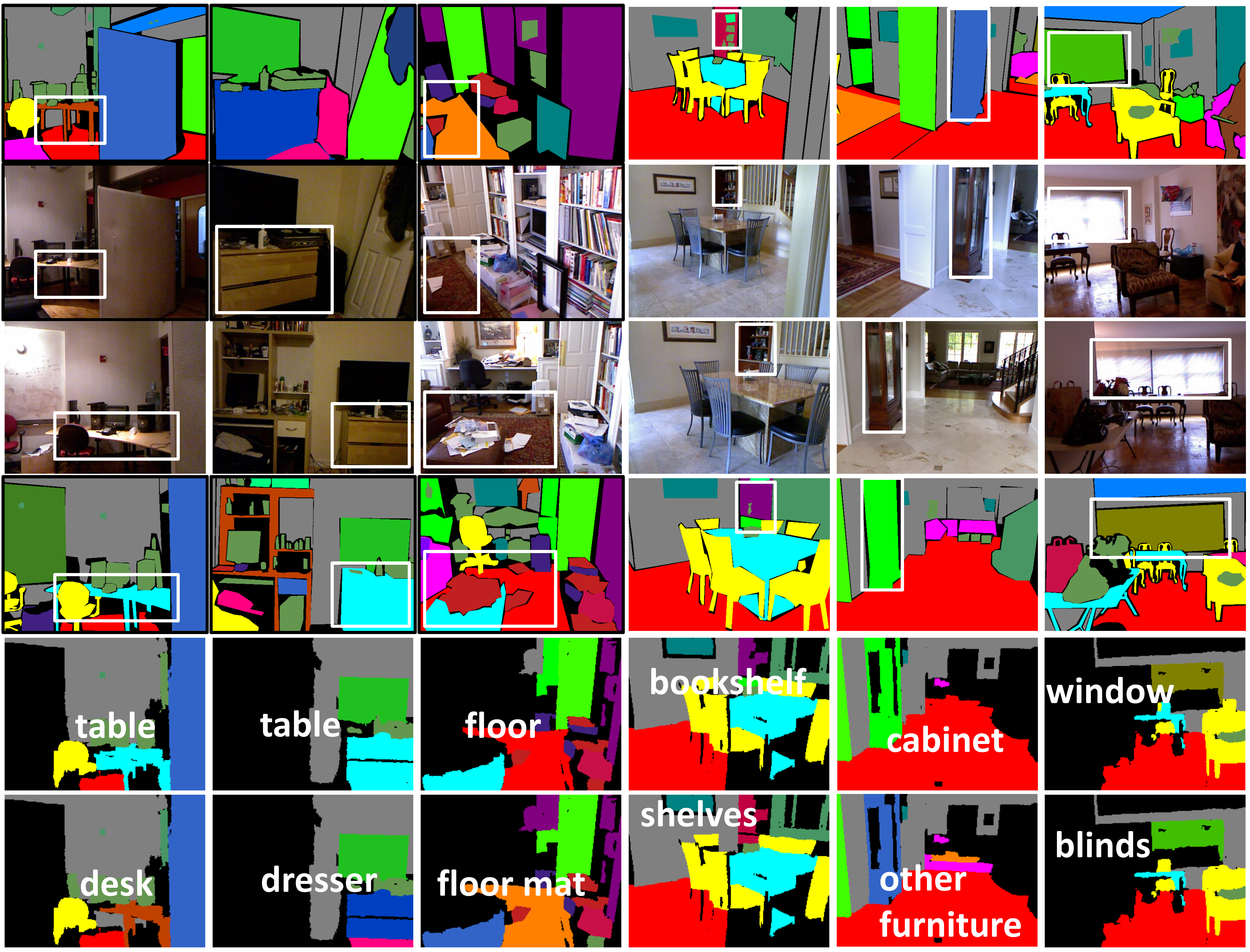}
\end{center}
   \caption{Example of groundtruth limitation and segmentation results of oracle case.
   Row 3 and 2 draw color images of target frame and next labeled frame, respectively. And row 4 and 1 draw their groundtruth.
   The segmentation result with groundtruth of target frame is shown in row 5, and the result with groundtruth of next frame is shown in row 6.
   We point out the regions in different frames with white bounding box, which are the same object of different views but labeled as different classes. }
\label{fig:gt}
\end{figure*}

At a closer look, it turns out that at least part of the performance loss in the best-case analysis for the correspondence is not due to bad matches between regions.
In Fig. \ref{fig:gt}, we present some examples of the annotations provided in the dataset.
In several cases, as the ones shown in the figure, the labeling is inconsistent and object labels are changed during the sequence.
From left to right in Fig. \ref{fig:gt}, table changes to desk, table changes to dresser, floor changes to floor mat, bookshelf changes to shelves, cabinet changes to other-furniture, and window changes to blinds.
Consequently, we see mistakes in the last two rows corresponding to the best case results due to inconsistent labelings.

\subsection{Qualitative results on NYUDv2 4-class and 13-class task}
\begin{figure*}[t]
\begin{center}
   \includegraphics[width=0.9\linewidth]{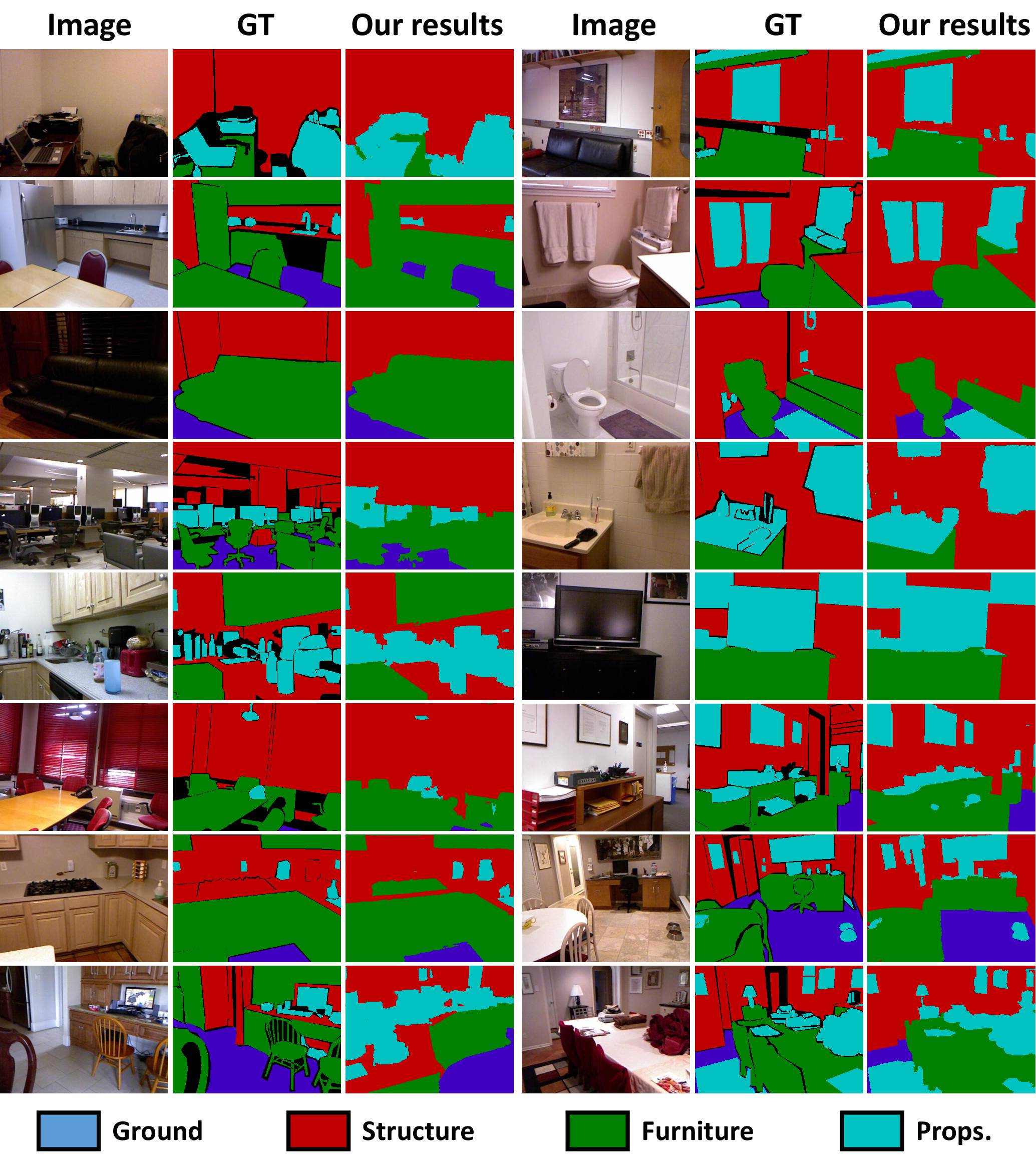}
\end{center}
   \caption{Semantic segmentation results of 4-class task on NYUDv2.}
\label{fig:4class}
\end{figure*}

\begin{figure*}[t]
\begin{center}
   \includegraphics[width=0.9\linewidth]{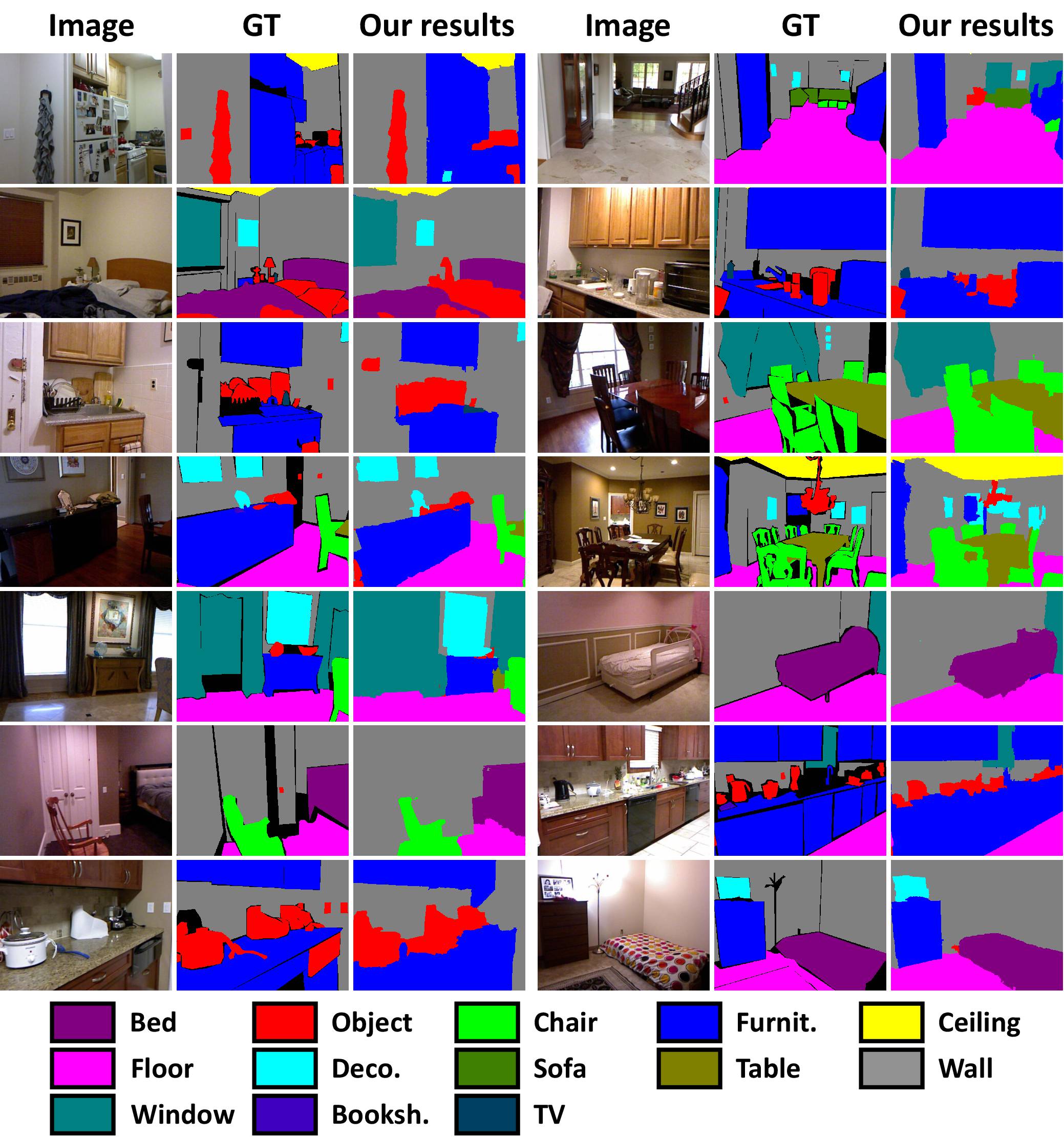}
\end{center}
   \caption{Semantic segmentation results of 13-class task on NYUDv2.}
\label{fig:13class}
\end{figure*}

We provide the qualitative results of 4-class and 13-class tasks of NYUDv2 dataset in Figure \ref{fig:4class} and Figure \ref{fig:13class} respectively.

\subsection{Qualitative results on NYUDv2 40-class task}
\begin{figure*}[t]
\begin{center}
   \includegraphics[width=0.95\linewidth]{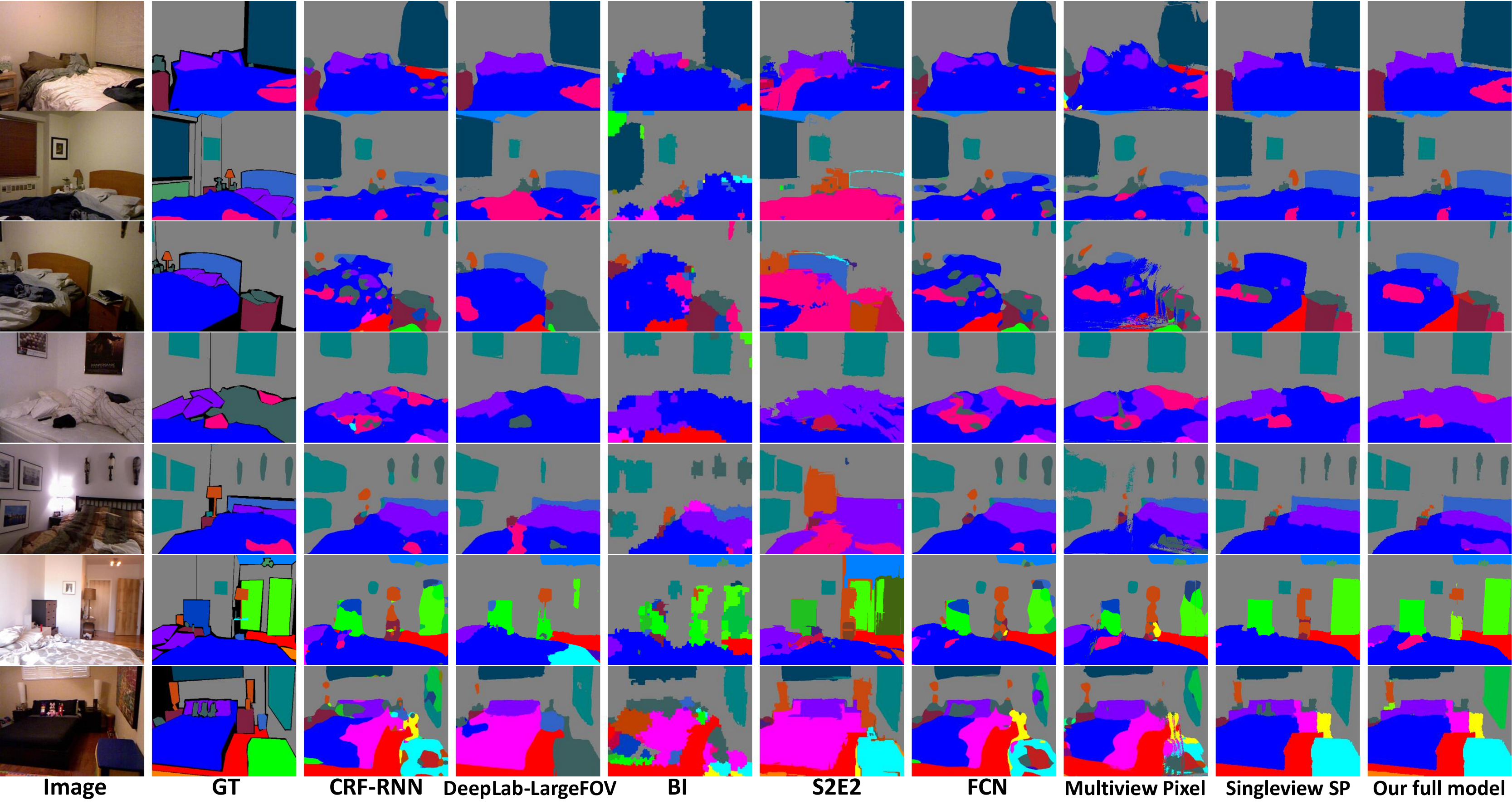}
\end{center}
   \caption{Semantic segmentation results of bedroom scenes on NYUDv2.}
\label{fig:bedroom}
\end{figure*}

\begin{figure*}[t]
\begin{center}
   \includegraphics[width=0.95\linewidth]{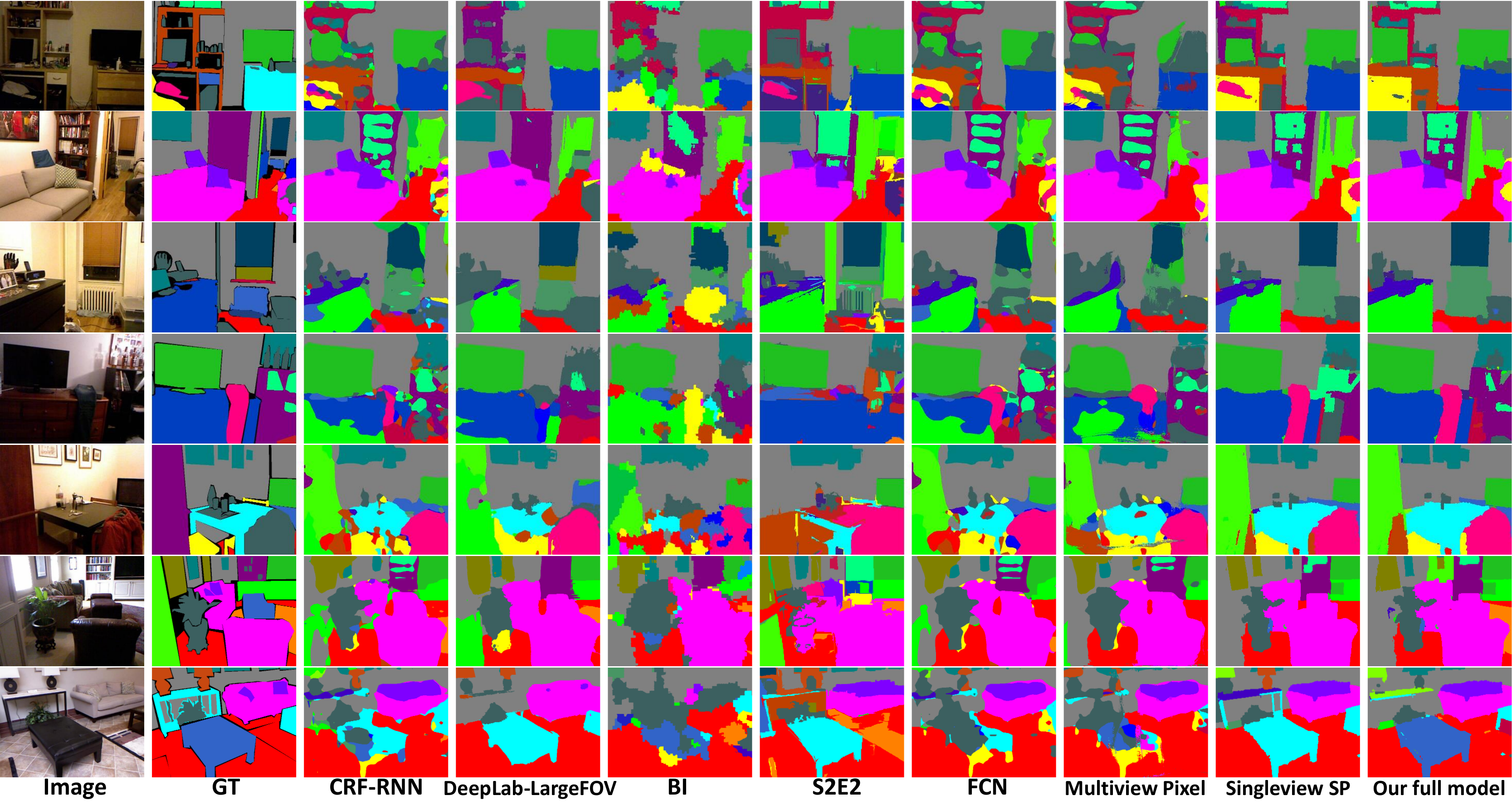}
\end{center}
   \caption{Semantic segmentation results of living room scenes on NYUDv2.}
\label{fig:living_room}
\end{figure*}

\begin{figure*}[t]
\begin{center}
   \includegraphics[width=0.95\linewidth]{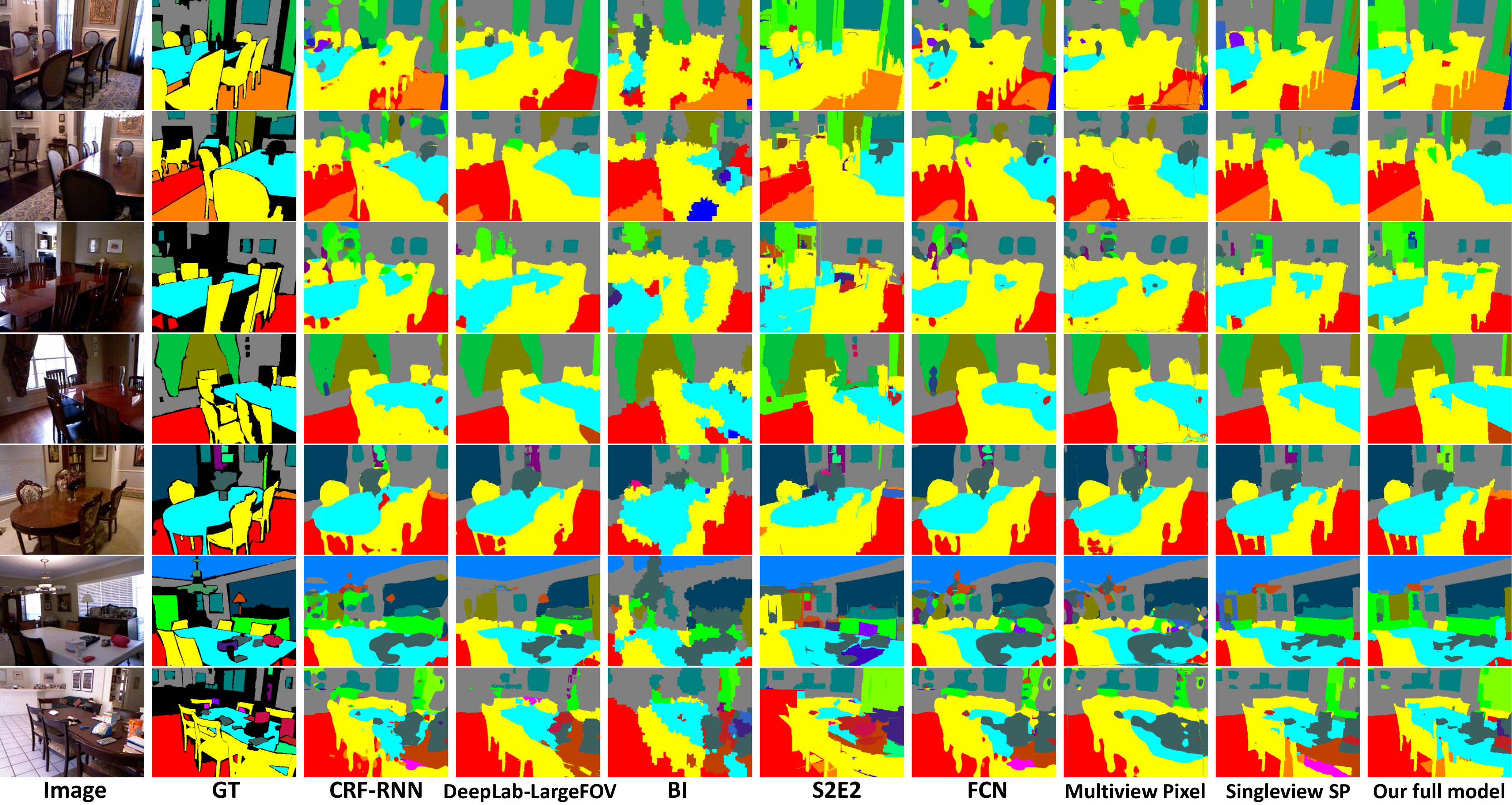}
\end{center}
   \caption{Semantic segmentation results of dining room scenes on NYUDv2.}
\label{fig:dining_room}
\end{figure*}

\begin{figure*}[t]
\begin{center}
   \includegraphics[width=0.95\linewidth]{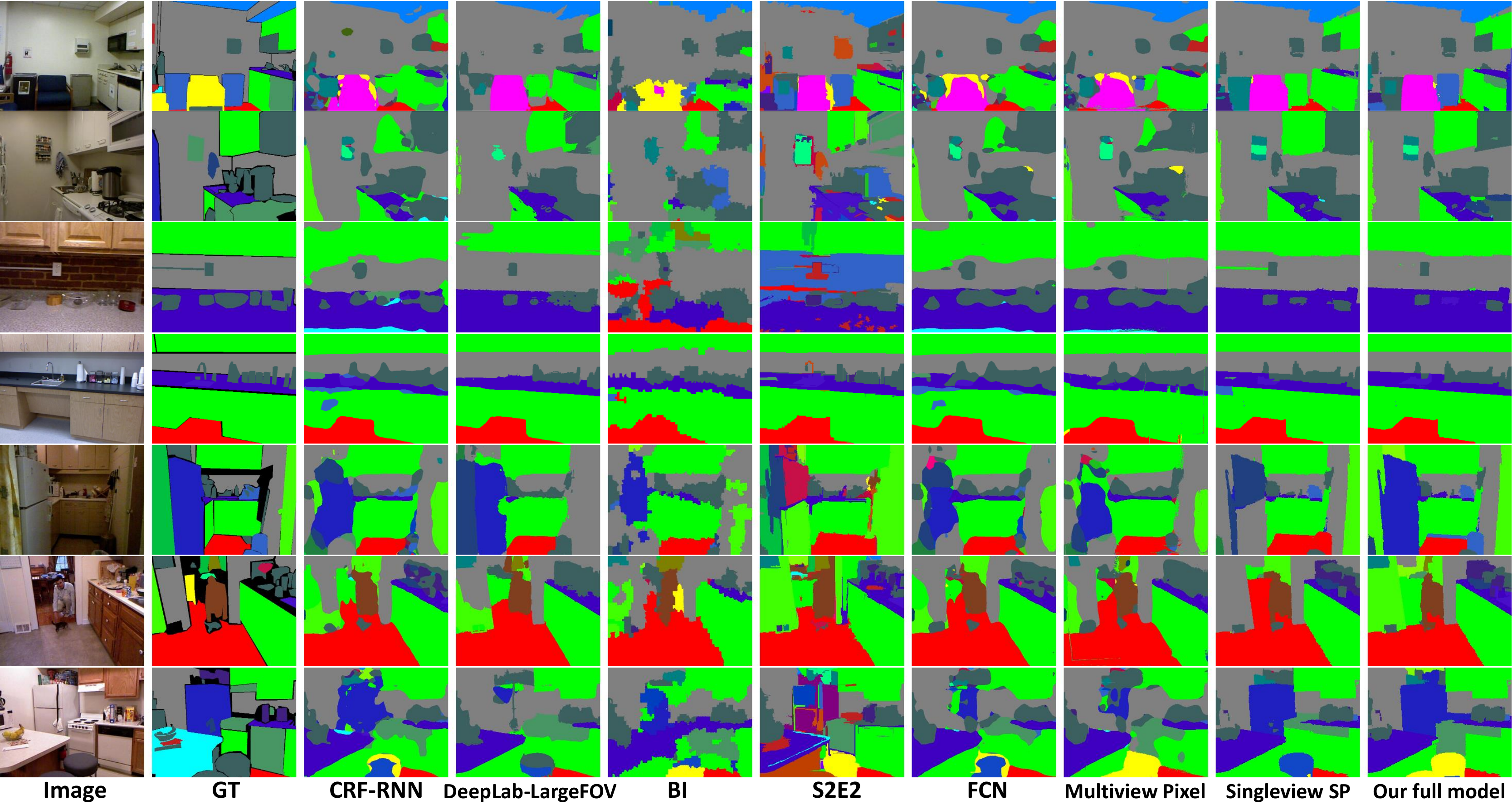}
\end{center}
   \caption{Semantic segmentation results of kitchen scenes on NYUDv2.}
\label{fig:kitchen}
\end{figure*}

\begin{figure*}[t]
\begin{center}
   \includegraphics[width=0.95\linewidth]{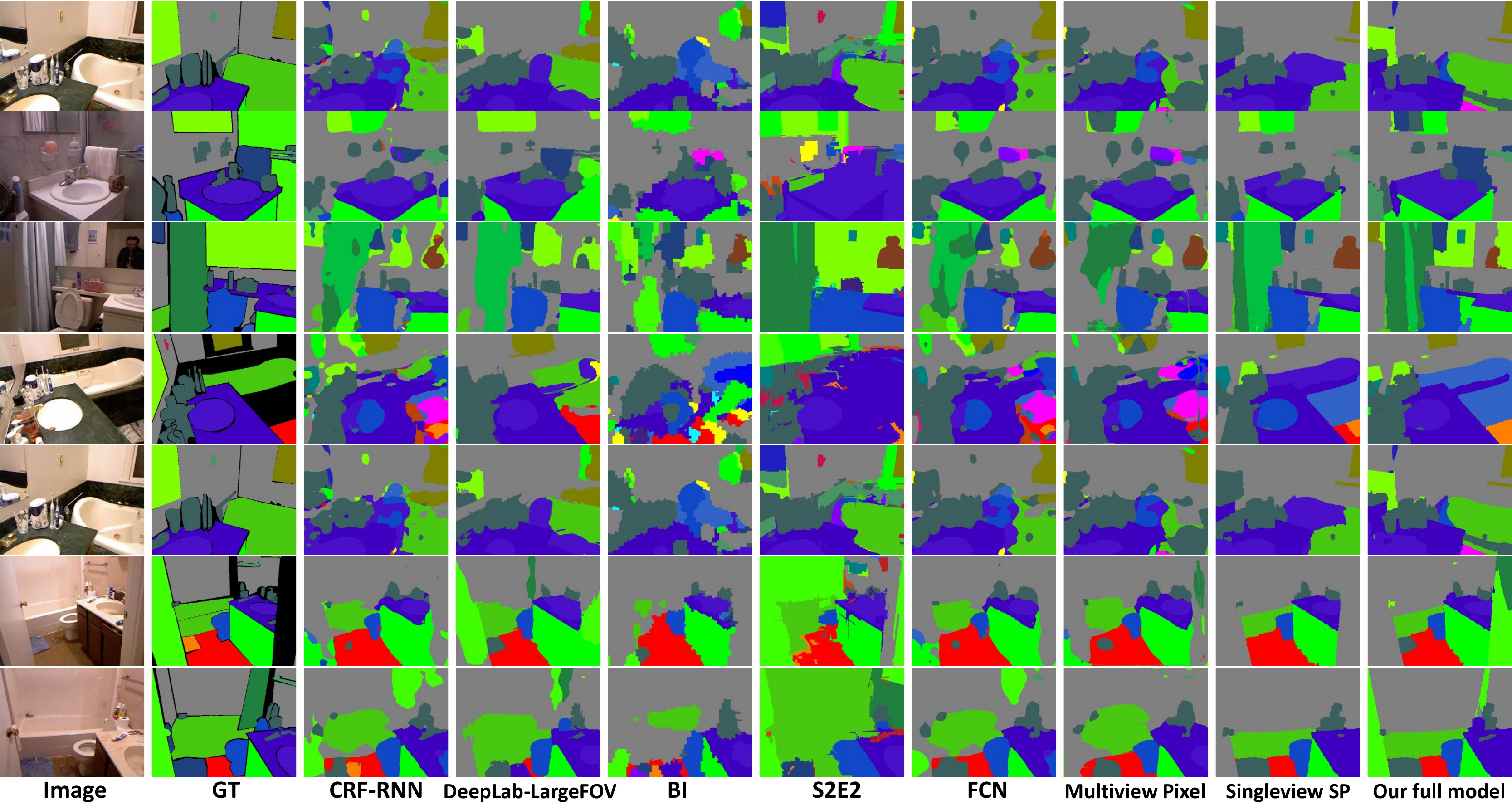}
\end{center}
   \caption{Semantic segmentation results of bathroom scenes on NYUDv2.}
\label{fig:bathroom}
\end{figure*}

\begin{figure*}[t]
\begin{center}
   \includegraphics[width=0.95\linewidth]{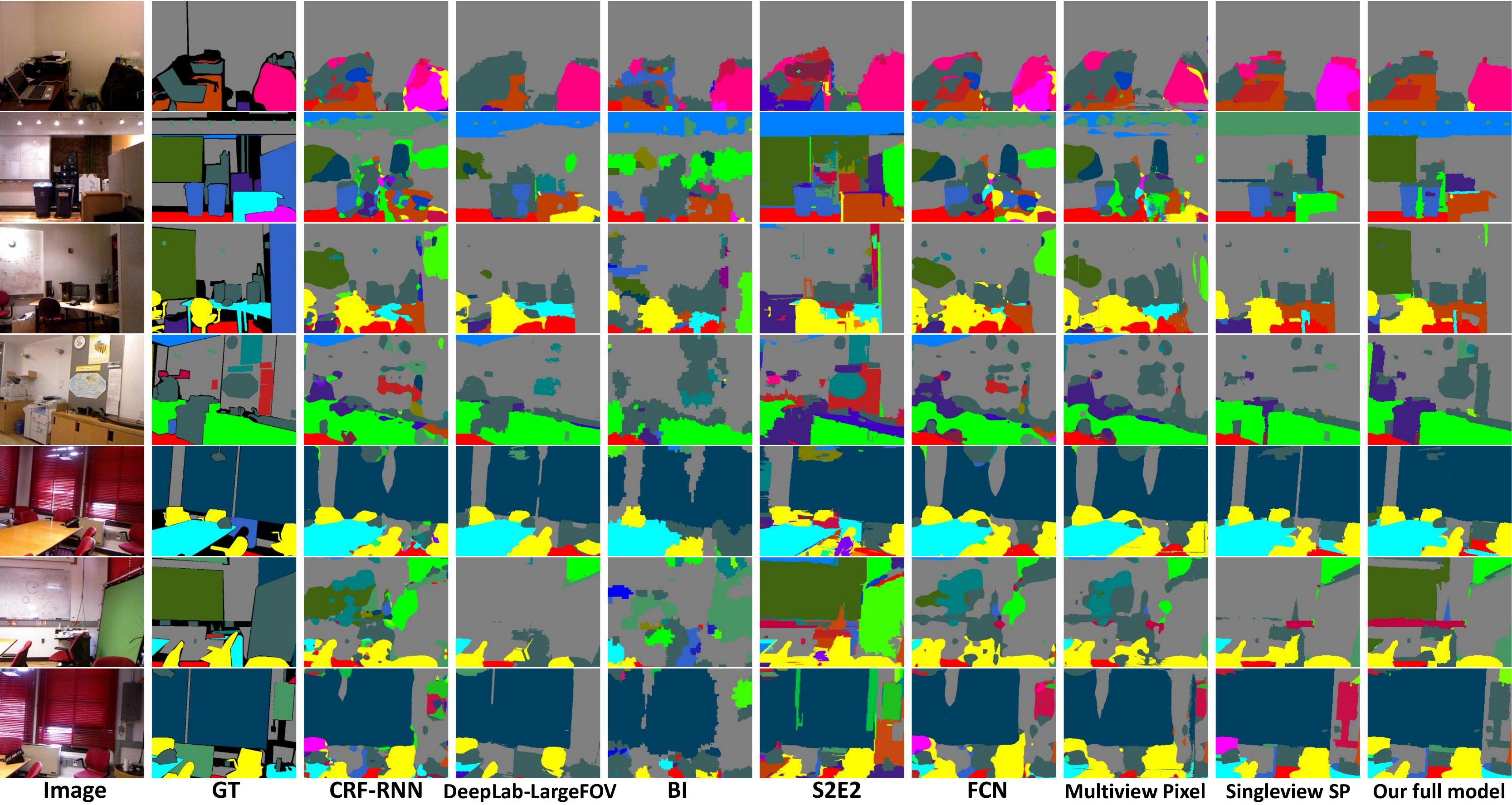}
\end{center}
   \caption{Semantic segmentation results of office scenes on NYUDv2.}
\label{fig:office}
\end{figure*}

\begin{figure*}[!t]
\begin{center}
   \includegraphics[width=0.95\linewidth]{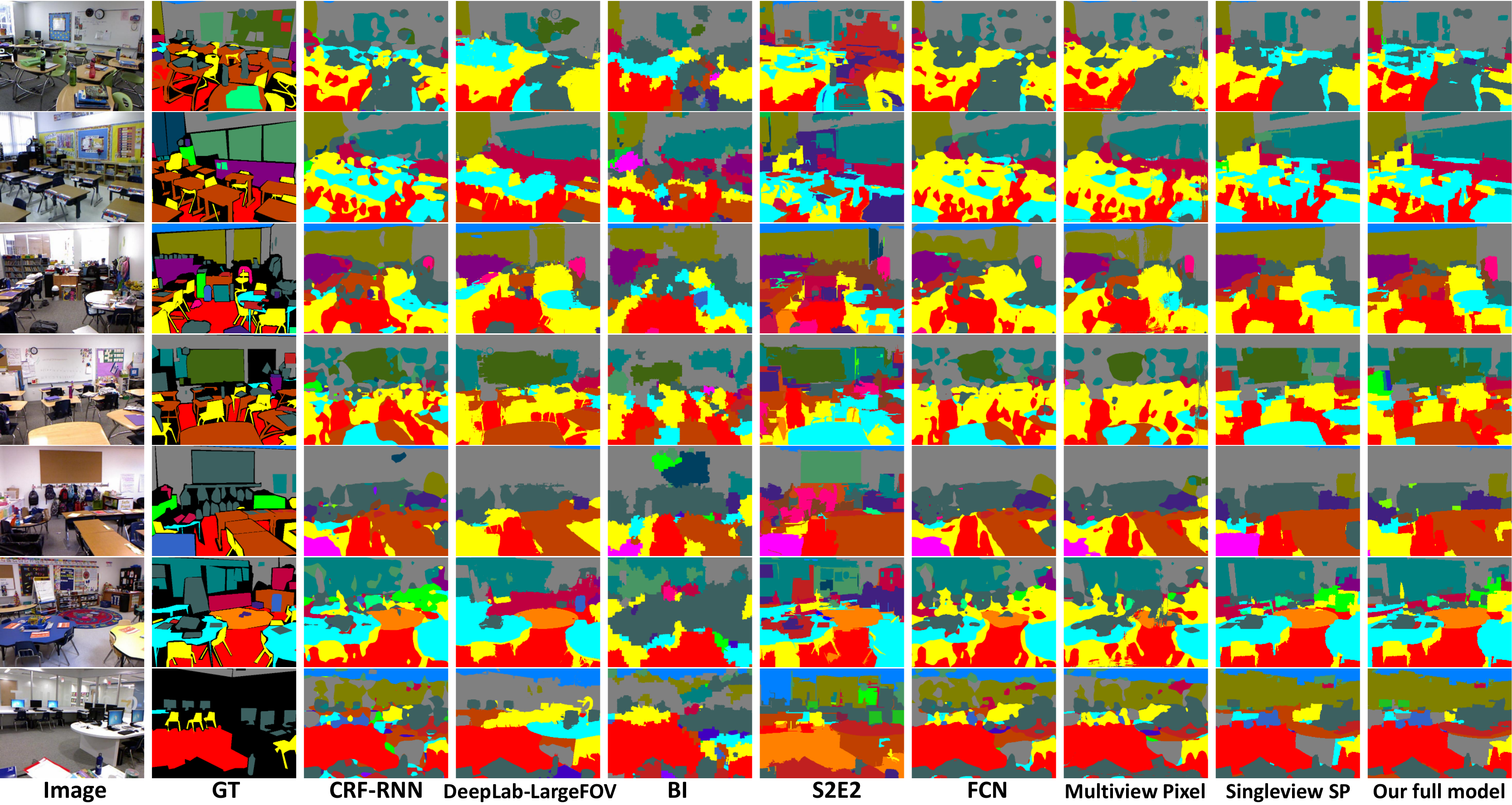}
\end{center}
   \caption{Semantic segmentation results of classroom scenes on NYUDv2.}
\label{fig:classroom}
\end{figure*}

We provide more qualititative results in the following figures. We pick up some
major scene categories from the test set including bedroom (Figure \ref{fig:bedroom}), living room (Figure \ref{fig:living_room}), dining room (Figure \ref{fig:dining_room}),
kitchen (Figure \ref{fig:kitchen}), bathroom (Figure \ref{fig:bathroom}), office (Figure \ref{fig:office}) and classroom (Figure \ref{fig:classroom}).

\subsection{Failure cases}
\begin{figure*}[t]
\begin{center}
   \includegraphics[width=0.7\linewidth]{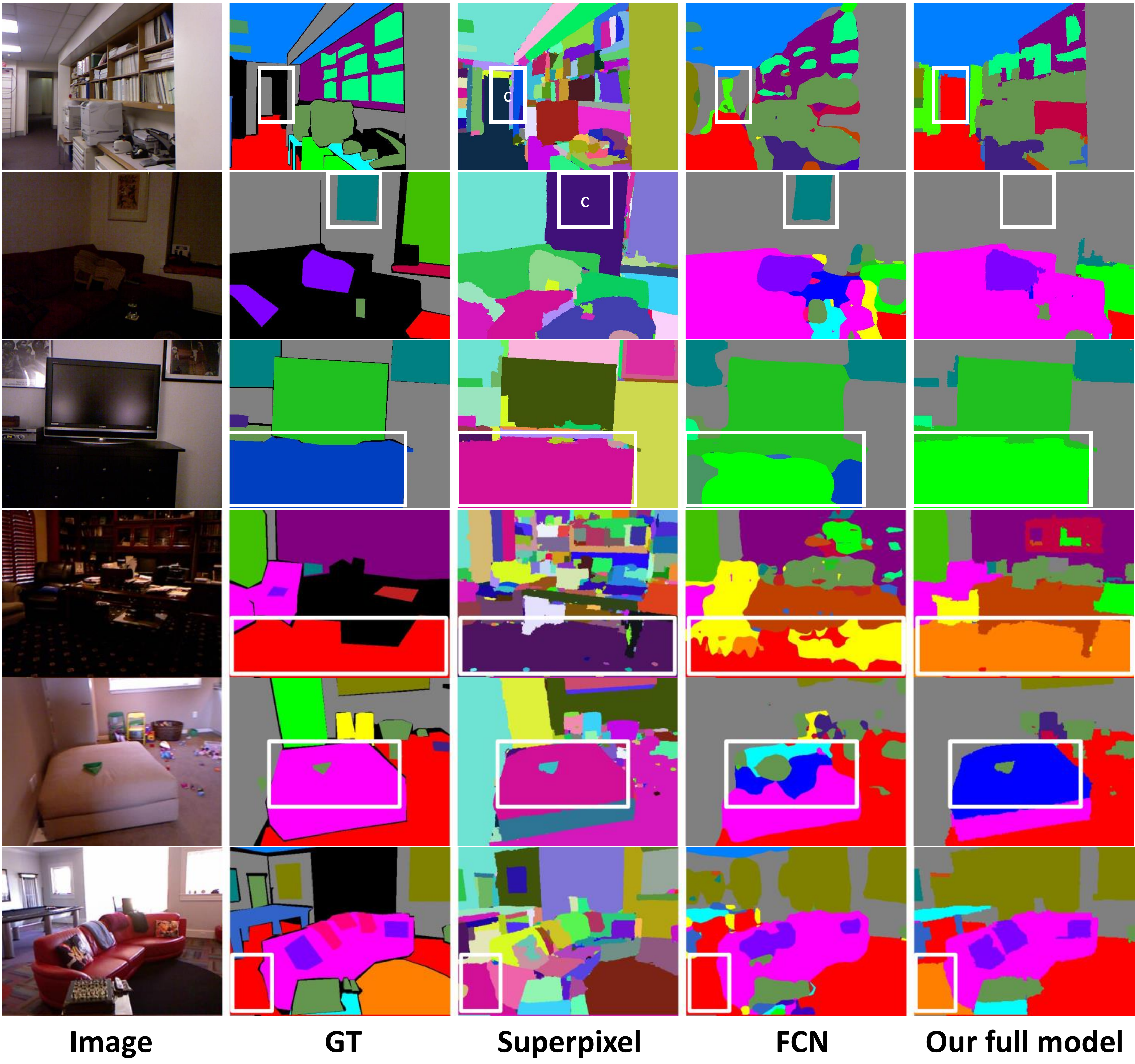}
\end{center}
   \caption{Some failure cases that our method is not able to improve FCN.}
\label{fig:failure}
\end{figure*}

In this section, we present some failure cases of our methods in Figure \ref{fig:failure}. In those views, our method does not achieve better result.
In the first two rows, we cannot segment the regions marked with white bounding box. This is because the superpixel in this two views cannot successful segment the regions.
We use the same parameter for all views, so it fails to provide good superpixels for our system, but we believe that it is not difficult to get better superpixels
for those failure views by adjusting the parameter of superpixel. In the third and fourth rows, we recognize the region as ``cabinet'' and ``floormat'' while groundtruth
are ``dresser'' and ``floor'', which are also difficult for human beings to classify.
In the last two rows, we show some challenges, which make our system fail to correctly recognize the region.

\end{document}